\documentclass{article}

\usepackage[final]{neurips_2025}

\usepackage[utf8]{inputenc} % allow utf-8 input
\usepackage[T1]{fontenc}    % use 8-bit T1 fonts
\usepackage{hyperref}       % hyperlinks
\usepackage{url}            % simple URL typesetting
\usepackage{booktabs}       % professional-quality tables
\usepackage{amsfonts}       % blackboard math symbols
\usepackage{nicefrac}       % compact symbols for 1/2, etc.
\usepackage{microtype}      % microtypography
\usepackage{xcolor}         % colors
\usepackage{graphicx} % Add this line to import the necessary package for including graphics

\usepackage{amsmath}
\usepackage{subfigure}
\usepackage{caption}

\usepackage{wrapfig}
\usepackage[toc,page,title]{appendix}
\usepackage{minitoc}
\usepackage[noend]{algpseudocode}
\usepackage{algorithm}
\usepackage{booktabs}

\usepackage{amsthm}

\newcommand{\tool}{Cago}

\DeclareMathOperator{\TV}{\mathbb{D}_{\text{TV}}}
\DeclareMathOperator{\E}{\mathbb{E}}

\title{Learning from Demonstrations via Capability-Aware Goal Sampling}

\author{%
  Yuanlin Duan\\
  Rutgers University\\
  \texttt{yuanlin.duan@rutgers.edu} \\
  \And
  Yuning Wang \\
  Rutgers University \\
  \texttt{yw895@cs.rutgers.edu} \\
  \AND
  Wenjie Qiu \\
  Rutgers University \\
  \texttt{wq37@cs.rutgers.edu} \\
  \And
  He Zhu \\
  Rutgers University \\
  \texttt{hz375@cs.rutgers.edu} \\
}

\begin{document}
\include{pythonlisting}

\maketitle

\begin{abstract} 

Despite its promise, imitation learning often fails in long-horizon environments where perfect replication of demonstrations is unrealistic and small errors can accumulate catastrophically. We introduce \tool{} (Capability-Aware Goal Sampling), a novel learning-from-demonstrations method that mitigates the brittle dependence on expert trajectories for direct imitation. Unlike prior methods that rely on demonstrations only for policy initialization or reward shaping, \tool{} dynamically tracks the agent's competence along expert trajectories and uses this signal to select intermediate steps—goals that are just beyond the agent's current reach—to guide learning. This results in an adaptive curriculum that enables steady progress toward solving the full task. Empirical results demonstrate that \tool{} significantly improves sample efficiency and final performance across a range of sparse-reward, goal-conditioned tasks, consistently outperforming existing learning from-demonstrations baselines.

\end{abstract}

\section{Introduction}
\label{Sec:Intro}

Imitation Learning (IL) provides a powerful paradigm for training agents using expert demonstrations, effectively alleviating the exploration challenges common in Deep Reinforcement Learning (DRL)~\citep{arulkumaran2024pragmatic}. The simplest form of IL is Behavior Cloning (BC), which directly supervises policy actions based on the states visited by the expert~\citep{bain1995framework, torabi2018behavioral}. However, BC often suffers from compounding errors when the learned policy deviates from expert trajectories. To overcome this limitation, modern approaches such as GAIL~\citep{ho2016generative}, PWIL~\citep{dadashi2020primal}, and AdRIL~\citep{eysenbach2021replacing} seek to align the state–action distributions of the agent and the expert through adversarial or distribution-matching objectives. In parallel, Inverse Reinforcement Learning (IRL) methods~\citep{ZiebartMBD08} aim to infer underlying reward functions from demonstrations, which can then guide reinforcement learning. More recently, advances in offline and offline-to-online RL, such as CQL~\citep{kumar2020conservative} and Cal-QL~\citep{nakamoto2023cal}, integrate demonstrations as anchors to regularize policy learning. These methods penalize value estimates that diverge from demonstrated behavior, mitigating overestimation and instability caused by out-of-distribution actions.

However, existing IL methods often struggle with complex, long-horizon tasks because they fail to reason about which parts of the task the agent has already mastered and which remain challenging. In particular, distribution-matching approaches perform flat matching—attempting to align occupancy measures over the entire trajectory distribution without considering the agent's evolving capabilities. This leads to poor exploration guidance, especially in the early stages of training when the agent seldom reaches meaningful parts of the state space. As a result, the learned reward function tends to assign uniformly low rewards, yielding uninformative gradients and hindering effective policy improvement. Some prior work proposes demonstration-guided curriculum learning that trains agents to solve tasks by starting near the goal or high-reward states and gradually expanding to earlier parts of the trajectories~\citep{Resnick2018,Salimans2018,TaoSC024}. However, these approaches rely on the ability to reset the agent to arbitrary demonstration states—an assumption impractical in real-world settings due to challenges in replicating physical conditions like joint velocities and angular momentum.

% In this paper, we propose a novel approach to leveraging demonstrations in DRL by focusing on guiding online trajectory sampling, thereby improving the quality of online data stored in the replay buffer. Specifically, we progressively guide the agent toward observations from demonstrations that lie at the boundary of its current capability, ensuring that the agent samples higher-quality trajectories. Since an agent’s initial capability is too weak to complete the full task, directly using the final task goal region as the agent’s objective would lead to poor trajectory sampling. Instead, we dynamically adjust the sampled goals based on the agent’s progress in replicating demonstrated trajectories, ensuring that the selected goals align with the agent’s learning stage. Our approach is evaluated in the context of Goal-Conditioned RL~\citep{liu2022goal, plappert2018multi}, where we could set observations from demonstrations as sampling targets.

We propose \tool{} (Capability-Aware Goal Sampling), a new learning-from-demonstrations framework that explicitly aligns the agent's learning process with its evolving capabilities. Unlike prior methods that use demonstrations for direct imitation, reward shaping, or offline pretraining, \tool{} treats demonstrations as structured roadmaps. It continuously monitors which parts of a demonstration the agent can already reach and leverages this signal to sample intermediate goal states in the demonstration, those at the boudary of the agent's current goal-reaching capabilities. At each episode, a goal-conditioned agent~\citep{liu2022goal, plappert2018multi} first attempts to reach the sampled goal and then explores forward from it, generating informative, task-relevant data for policy optimization. This iterative process of capability-aware goal selection and curriculum-aligned exploration enables steadily progress toward solving the full task. 

%To make this curriculum practical in real-world settings, \tool{} resets episodes only to the initial states of demonstration trajectories—an operation that is far more feasible than initializing from arbitrary intermediate points. 

We evaluate \tool{} across several sparse-reward environments and demonstrate substantial improvements in both sample efficiency and final task performance over existing imitation-based baselines. Our experiments highlight that capability-aware goal sampling provides a powerful signal for structuring learning, particularly in long-horizon tasks.

\section{Background and Problem Setup}
\label{Sec:PSB}

\textbf{Reinforcement learning} (RL) aims to enable agents to learn optimal behaviors through trial-and-error interactions with an environment. An RL problem is formulated as a Markov Decision Process (MDP), represented as a tuple \((\mathcal{S}, \mathcal{A}, \mathcal{T}, \mathcal{G}, \eta, R, \rho_0)\). The agent operates within a state space \(\mathcal{S}\) and takes actions from an action space \(\mathcal{A}\), transitioning between states according to the dynamics \(\mathcal{T}(s'|s, a)\). $R(s,a) \in \mathbb{R}$ is the reward function and $\rho_0$ is the initial state distribution. Given a policy $\pi$, consider the trajectory $\tau = \{s_0, a_0, s_1, a_1, \ldots\}$ sampled by $\pi$, i.e., $s_0 \sim \rho_0$, $a_t \sim \pi(\cdot | s_t)$, and $s_{t+1} \sim \mathcal{T}(\cdot | s_t, a_t)$. The goal of RL is to learn a return-maximizing policy $\pi^* = \arg\max_\pi \mathbb{E}_{\tau \sim \pi(a_t \mid s_t)} \left[ \sum_{t=0}^{\infty} \gamma^t r(s_t, a_t) \right]$ where $\gamma \in (0, 1]$ is the discount factor.

\textbf{Learning from Demonstrations.} In imitation learning, the agent is provided a dataset of demonstrations $\mathcal{D}_\text{demo}$ collected from some expert policy $\pi_\text{expert}$. The objective is to learn a policy $\pi$ that reproduces the expert’s behavior by generalizing from these demonstrations. The simplest approach, behavioral cloning (BC), treats this as a supervised learning problem, minimizing the discrepancy between the agent’s predicted actions and the expert’s.
Another line of work, inverse reinforcement learning (IRL)~\citep{PieterN04}, aims to infer an underlying reward function that explains the expert’s behavior and then optimizes a policy through RL on this learned reward, thus decoupling reward inference from policy optimization.
Building on ideas from IRL, Generative Adversarial Imitation Learning (GAIL)~\citep{ho2016generative} bypasses explicit reward recovery by training a policy and a discriminator in an adversarial game: the discriminator distinguishes expert from agent trajectories, while its output serves as an implicit, learned reward signal guiding the policy.
More broadly, many recent imitation learning algorithms can be interpreted as minimizing a divergence between the expert and agent occupancy measures.

%The goal is to learn a policy $\pi$ that mimics the expert's behavior by generalizing from these demonstrations. A common approach is behavioral cloning, which casts the problem as supervised learning, training the agent to minimize the discrepancy between its predicted actions and the expert's. An alternative approach is to infer a reward function from expert demonstrations and then optimize a policy through RL on this learned reward. This class of methods, known as inverse RL (IRL)~\citep{PieterN04}, decouples the tasks of reward inference and policy learning. A notable example is Generative Adversarial Imitation Learning (GAIL)~\citep{ho2016generative}, which frames imitation as a minimax game where a discriminator distinguishes expert and agent behaviors, effectively serving as a learned reward signal for policy optimization. Many recent work in imitation learning can be interpreted through the lens of minimizing a divergence between the expert and agent occupancy measures.

\textbf{State Reset.} Several methods attempt to mitigate the exploration challenge in sparse-reward RL environments by resetting the agent to states from expert demonstrations, thereby bypassing the need to discover those states through the agent's own exploration. These strategies include initializing the agent to states sampled uniformly from demonstration trajectories~\citep{nair2018overcoming,PengALP18,HosuR16}, employing a hand-crafted curriculum~\citep{Zhu0MRECTKHFH18}, or using a reverse curriculum that progressively trains the agent from goal or high-reward states backward~\citep{Resnick2018,Salimans2018,TaoSC024}. These approaches assume the ability to reset the agent to arbitrary demonstration states—an assumption that is unrealistic in real-world settings.

\textbf{Goal-Conditioned RL} %A fundamental challenge in RL is efficiently achieving long-horizon goals, especially when reward signals are sparse. 
%Goal-conditioned reinforcement learning 
(GCRL) extends the standard RL framework by conditioning policies on specific target goals, guiding agents toward desired goals. The MDPs are augmented with a goal space \(\mathcal{G}\) and are associated with states via a mapping \(\eta : \mathcal{S} \to \mathcal{G}\), ensuring that each state corresponds to an achieved goal. In GCRL, the reward signal from the environment is typically sparse and is defined as: \(R(s, a, s', g) = 1\{\eta(s') = g\}\).
We assume that each episode has a fixed horizon \(T\) and $\mathcal{S} = \mathcal{G}$. The agent's objective is to train a goal-conditioned policy $\pi^G(\cdot | s_t, g)$ to achieve a given goal \(g \in \mathcal{G}\) through maximizing the expected cumulative reward \(J(\pi) = \mathbb{E}_{g \sim p_g, \tau \sim \pi^G(a_t \mid s_t, g)} \left[ \sum_{t=0}^{T-1} \gamma^t \cdot R(s_t, a_t, s_{t+1}, g) \right]\) where \(p_g\) is the goal distribution.

\textbf{This Paper.} We introduce \tool{}, a novel approach that leverages demonstrations as a scaffold for goal-directed reinforcement learning. Rather than direct imitation, \tool{} uses demonstrations to guide exploration by training a goal-conditioned policy $\pi^G(a \mid s, g)$ that learns to progressively reach intermediate states $g$ along demonstration trajectories, effectively inducing a curriculum that facilitates steady progress toward solving the full task. In addition, \tool{} learns a goal predictor $\mathcal{P}(s)$ that infers the final goal state $g_T$ from the current state $s$. The resulting task policy is defined as $\pi(s) = \pi^G(s, \mathcal{P}(s))$ that enables automatic inference of goal conditions at test time for previously unseen situations.
%enabling the agent to generalize to novel situations by reaching inferred goals.

\section{Method}
\label{Sec:Method}

The main idea of \tool{} is to continuously monitor the agent's evolving capabilities to reach various stages of demonstration trajectories during training. It dynamically selects the most appropriate goal from the demonstrations, conditioned on the agent's current performance ceiling. The selected goal guides online exploration, with the agent first attempting to reach it using its current policy. From there, it continues to explore, collecting task-relevant trajectories in an Go-Explore style~\citep{ecoffet2019go}. By anchoring exploration in achievable yet progressively harder goals, this process effectively constructs an implicit curriculum, where the agent is gradually exposed to more challenging states aligned with its growing competence. 
%By aligning goal selection with the agent's current capability, \tool{} ensures that sampled trajectories reflect meaningful progress toward task completion, leading to data that is both valuable for learning and relevant to the task. 
%Critically, by selecting goals that are just beyond the agent’s present abilities, we avoid the pitfalls of unproductive exploration caused by overly ambitious targets, while still challenging the agent to improve. This strategy promotes the collection of higher-quality data, which in turn facilitates the training of more accurate world models and stronger policies—ultimately resulting in substantially improved learning efficiency and results.

\subsection{Observation Visit Tracking with Demonstration Alignment}

\tool{} assumes the existence of a limited amount of expert demonstrations
%Our approach circumvents these limitations by leveraging minimal demonstration trajectories for evaluating agent capabilities and goal sampling. Suppose we have a small demonstration dataset 
$\mathcal{D}_{\text{demo}}: \{\tau^{(i)} = \{(s_0, a_0)^{(i)}, \dots, (s_{L_i},a_{L_i})^{(i)}\}\}_{i=1}^M$ where $M$ is the number of the demonstrations and $L_i$ is the length of $i$-th demonstration $\tau^{(i)}$.
To select goals at the boundary of the agent's current reaching capabilities,
%that challenge the agent without exceeding its upper capability limit, thereby enhancing the quality of online data, 
It is crucial to determine the stage at which the agent can accomplish its task completion. Central to our method is maintaining a dictionary $\text{Dict}_{\text{visit}}$ that tracks the visitation frequencies of observations $s_i$ across demonstrations. 
%Assuming each demonstration is with a unique environment reset seed, our goal-conditioned policy samples trajectories by resetting the environment to the same seed of a randomly selected demonstration. This ensures consistent environmental conditions with the demonstration. 
For each demonstration $\tau^{(i)}$, we initialize an all-zero list of the same length as its steps. Each element in this list records the visitation count of the corresponding observation in the demonstration. At each environment step, the record list is updated to reflect whether the agent has visited observations from the demonstration, based on similarity metrics $\text{sim}(\cdot, \cdot)$ such as L2 distances for state-based environments or mean squared errors (MSE) between images in visual environments. Formally, we define the visitation record dictionary as: 
\begin{equation} \label{DictDefine}
\text{Dict}_{\text{visit}} = { \tau^{(i)} : [0, 0, \dots, 0] \in \mathbb{N}^{L_i} \mid i = 1, 2, \dots, M } 
\end{equation} 
%where $seed_i$ is the environment reset seed corresponding to the $i$-th demonstration $\tau^{(i)}$.
During online exploration, for each new episode, we first sample a demonstration $\tau^{(i)}$ from $\mathcal{D}_{\text{demo}}$ and reset the environment to the initial state of $\tau^{(i)}$. \textbf{Our strategy is more practical in the real-world setting} than related methods (e.g. \citep{TaoSC024,nair2018overcoming}) that reset the environment to intermediate demonstration states, which are often infeasible to reproduce due to unobservable or difficult-to-control physical factors such as velocity and angular momentum.
%for a trajectory rollout under seed $seed_i$, 
Given a rollout $\tau = (s_0, s_1, \ldots)$ from the environment, we update $\text{Dict}_{\text{visit}}[\tau^{(i)}]$ as follows: 
\begin{equation} \label{DictUpdate}
\text{Dict}_{\text{visit}}[\tau^{(i)}][j] \mathrel{+}= 1 \quad \text{if } \text{sim}(s_t, s^{(i)}_j) \leq \epsilon, \quad \forall t \in {1, \dots, L_{\tau}}, \forall j \in {1, \dots, L_i} 
\end{equation} 
where $s_t$ is the agent’s observation state at timestep $t$, $L_{\tau}$ is the total length of the rollout $\tau$, $s^{(i)}_j$ is the $j$-th observation state in the $i$-th demonstration $\tau^{(i)}$, $\text{sim}(\cdot, \cdot)$ is the similarity metric (e.g., L2 distance for state-based environments or MSE for image-based environments), and $\epsilon$ is a matching threshold. This simple record dictionary effectively tracks the agent's progress and helps identify its goal-reaching capability limits along task demonstrations. %enabling the selection of appropriate goals.

%This straightforward record dictionary proves invaluable for gauging the goal-conditioned policy's proficiency in reaching various stages of task completion. It also provides insights into the agent's current capacity limits for task completion, thereby enabling the establishment of suitable goals for the agent.

% \begin{algorithm}[h]
% \caption{Capability-Aware Goal Sampling}
% \label{Algorithm-FindGoal}
% \begin{algorithmic}[1]
%     \State \textbf{Input:} Demonstration trajectory $Demo_{use}$, Environment seed $seed$, Demo visit record dict $\text{Dict}_{\text{visit}}$
%     \State \textbf{Output:} Selected goal $g$
%     \State Length $l = |Demo_{use}|$
%     \State $index \leftarrow$ Find last index where $\text{Dict}_{\text{visit}}[seed] \geq threshold\ \ \lambda_{visit}$(e.g. 100)
    
%     \State $min\_index = \max(0, index - \frac{l}{10})$
%     \State $max\_index = \min(l, index + \frac{l}{10})$
    
%     \State $g$ $\leftarrow$ Randomly sample a observation from $Demo_{use}$ in index range $[min\_index, max\_index]$
%     \State
%     \Return $g$
% \end{algorithmic}
% \end{algorithm}

\subsection{Capability-Aware Goal Sampling}

\begin{figure}[t]
    \centering
    \includegraphics[width=1\textwidth]{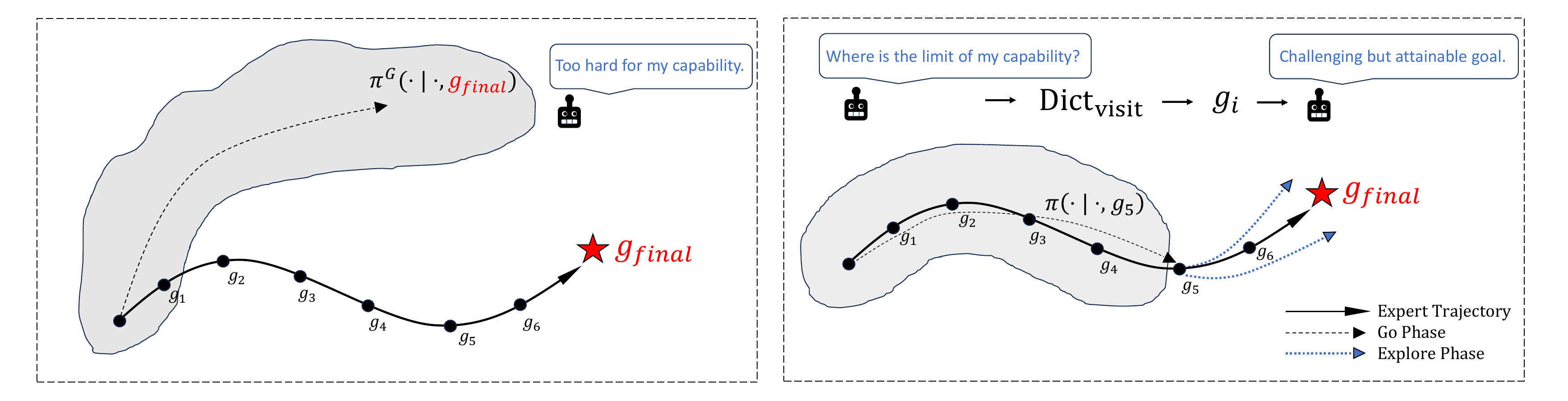}
    \caption{Illustration of the \tool{}. 
        \textbf{Left:} Directly setting the final goal as the agent’s target often leads to failure, as the current policy $\pi^G$ may not yet be capable of reaching it. The shaded region illustrates the set of states currently reachable under $\pi^G$. Attempting to reach $g_{\text{final}}$ (i.e., executing $\pi^G(\cdot|\cdot, g_{\text{final}})$)  causes the agent to diverge from the demonstration trajectory.
        \textbf{Right:} \tool{} improves learning by leveraging a visitation frequency dictionary $\text{Dict}_{\text{visit}}$ built from demonstrations. Given a demonstration trajectory with subgoals $g_1, g_2, \dots, g_n$, the agent selects the furthest subgoal $g_i$ that remains within its current capabilities for Go-Explore sampling, enabling a curriculum of progressively more challenging goals aligned with the demonstration.
        %\tool{} improves learning by leveraging a visitation frequency dictionary $\text{Dict}_{\text{visit}}$ constructed from demonstrations. Given a demonstration with subgoal states $g_1, g_2, \dots, g_n$, the agent determines the last subgoal $g_i$ where the visitation frequency exceeds a threshold—representing the agent's current capability limit. %The agent samples a goal from a range centered around $g_i$, which is neither too easy nor too difficult, fostering progressive learning.
        }
    \label{fig:method}
\end{figure}

\tool{} leverages demonstration-based visitation counts to guide goal selection and trajectory collection in a \textit{capability-aware} manner. Figure~\ref{fig:method} illustrates our method. After resetting the environment to the initial state of a randomly sampled demonstration $\tau^{(i)}$, we select a goal $g$ for the agent to explore:
\begin{equation}
g \sim \mathcal{G}_{\text{cap}}(\pi^G, \tau^{(i)}),
\end{equation}
where $\mathcal{G}_{cap}(\pi^G, \tau^{(i)})$ denotes a capability-aware goal sampling distribution over subgoals whose reachability is aligned with the current goal-reaching capability of the policy $\pi^G$. 
%Prior to sampling a new trajectory from the environment, the agent first selects a demonstration $\tau_{\text{expert}}^{i}$ and resets the environment using the same seed $seed_i$. Then, 
\tool{} examines the visitation frequency list to identify the last index where the frequency exceeds a predefined threshold:
\begin{equation}\label{capability-aware}
j^* = \max \left\{ j \mid \text{Dict}_{\text{visit}}[\tau^{(i)}][j] \geq \lambda_{\text{visit}} \right\},
\end{equation}

\begin{wrapfigure}[9]{r}{0.59\textwidth}
\vspace{-15pt}
\scalebox{0.8}{
\begin{minipage}{0.72\textwidth}
\begin{algorithm}[H]
\caption{Capability-Aware Goal Sampling (\tool{})}
\label{Algorithm-FindGoal}
\begin{algorithmic}[1]
    \State \textbf{Input:} Demonstration $\tau^{(i)}$, Visitation record $\text{Dict}_{\text{visit}}$
    \State \textbf{Output:} capability-aware goal $g$
    \State Identify capability-aware upper limit point $g_i$ using visitation threshold $\lambda_{\text{visit}}$ (Eq.~\eqref{capability-aware})
    \State Define sampling region $\mathcal{G}_{cap}(\pi^G, \tau^{(i)})$ centered around $g_i$ (Eq.~\eqref{samplerange})
    \State Sample subgoal $g \sim \mathcal{G}_{cap}(\pi^G, \tau^{(i)})$
    \State \Return $g$
\end{algorithmic}
\end{algorithm}
\end{minipage}
}
\end{wrapfigure}

where $\text{Dict}_{\text{visit}}[\tau^{(i)}][j]$ denotes the visitation frequency of $j$-th observation $s_j$ of $\tau^{(i)}$ under policy $\pi^G$ and $\lambda_{\text{visit}}$ is a frequency threshold (e.g. 100). This index indicates the latest point in the demonstration that the agent is sufficiently competent at reaching—effectively serving as a proxy for the limit of the agent's current goal-reaching capability. 
\tool{} constructs a goal sampling range centered around $j^*$. The sampled goal is drawn from this range, which allows the agent to either revisit familiar goals or attempt slightly more challenging ones that are just beyond its current capability. This capability-aware goal sampling strategy introduces controlled diversity into the training process and encourages progressive learning, while also avoiding excessively difficult goals that could derail training. The corresponding goal sampling region is defined as:
\begin{equation}\label{samplerange}
\mathcal{G}_{cap}(\pi^G, \tau^{(i)}) = \left\{ s_k \in \tau^{(i)} \mid |k - j^*| \leq \delta \cdot L_i \right\},
\end{equation}
where $L_i$ is the length of $\tau^{(i)}$ and $\delta \in (0,1]$ controls the window size for goal sampling (e.g., 10\% of the demonstration length). Goals are then sampled at random from this set. Our capability-aware goal sampling scheme introduces a curriculum-aligned learning signal that progressively guides the agent with steady improvement toward successful task completion.
%adaptively reducing the divergence between policy rollouts and expert behavior, thereby minimizing the adaptation error term in Equation~\eqref{returnbound} and improving policy performance in the real-world environment. 
%In practice, with a small probability, we include the final goal of the demonstration in the sampling process to remind the agent of the ultimate task objective. 
The overall goal sampling process in \tool{} is described in Algorithm~\ref{Algorithm-FindGoal}. 

% \begin{algorithm}[t]
% \caption{Capability-Aware Goal Sampling (\tool{})}
% \label{Algorithm-FindGoal}
% \begin{algorithmic}[1]
%     \State \textbf{Input:} Demonstration trajectory $\tau^{i}_{\text{expert}}$, Environment seed $s$, Visitation record $\text{Dict}_{\text{visit}}$
%     \State \textbf{Output:} goal $g$
%     \State Identify capability-aware upper limit point $g_i$ using visitation threshold $\lambda_{\text{visit}}$ (Eq.~\eqref{capability-aware})
%     \State Define sampling region $\mathcal{G}_{cap}(\pi)$ centered around $g_i$ (Eq.~\eqref{samplerange})
%     \State Sampled subgoal $g \sim \mathcal{G}_{cap}(\pi)$
%     \State \Return $g$
% \end{algorithmic}
% \end{algorithm}

\subsection{Learning Framework}
\label{subsec: framework}

\textbf{Go-Explore.}
% Our approach draws inspiration from the Go-Explore paradigm~\citep{ecoffet2019go}, which splits each training episode into two distinct phases: the Go-phase and the Explore-phase. In the Go-phase, the agent is guided toward the sampled goal $g$ using the GCRL policy $\pi^G$, reaching an intermediate state $s_{T}$. In the subsequent Explore-phase, an exploration policy $\pi^E$ takes over from $s_{T}$ for the remaining time steps, enabling broader environment interaction. To implement the Explore-phase, we propose a novel exploration strategy called the Behavior Clone (BC) Explorer. This policy is trained using only a limited number of demonstrations and produces a stochastic action distribution. During exploration, we sample actions from this distribution to balance between exploration and imitation. This allows the agent to explore in a goal-directed yet behaviorally plausible manner. We further analyze the impact of the BC Explorer through ablation studies, detailed in Section~\ref{Sec:Experiments}.
%With capability-aware goals in hand, we proceed to collect trajectories that facilitate both effective learning and broader exploration. 
\tool{} trains a goal-conditioned agent following the Go-Explore paradigm~\citep{ecoffet2019go}, which divides each episode into two sequential phases: the \textit{Go-phase} and the \textit{Explore-phase}. In the Go phase, the agent is guided towards a sampled goal state $g$ using the goal-conditioned policy $\pi^G(\cdot | \cdot, g)$, reaching an intermediate state $s_{E}$. To improve environment exploration beyond the agent's current capabilities, the Explore-phase takes over from $s_{E}$, where an exploration policy $\pi^E$ is used for the remaining time steps. Since we have access to a limited set of task demonstrations $\mathcal{D}_{\text{demo}}$, we implement $\pi^E$ as a Behavior Clone (BC) Explorer trained on  $\mathcal{D}_{\text{demo}}$. The BC Explorer outputs a stochastic action distribution that enables the agent to balance between exploration and imitation. This two-phase strategy ensures that collected trajectories stay anchored near the demonstration distribution in $\mathcal{D}_{\text{demo}}$, while encouraging exploration. We further analyze the impact of the BC Explorer through ablation studies, detailed in Section~\ref{Sec:Experiments}. 

As \tool{} actively resets environments to initial states drawn from the demonstration set $\mathcal{D}_{\text{demo}}$, a key question is how the agent generalizes beyond $\mathcal{D}_{\text{demo}}$. Our solution is to train the goal-conditioned agent $\pi^G$ using a richer set of imagined rollouts generated by a world model via model-based RL.

\begin{wrapfigure}[19]{r}{0.59\textwidth}
\vspace{-15pt}
\scalebox{0.8}{
\begin{minipage}{0.72\textwidth}
\begin{algorithm}[H]
\caption{The main training framework of \tool{}}
\label{Algorithm-TrainingFrame}
\begin{algorithmic}[1]
    \State \textbf{Input:} GC Policy $\pi^G$, World Model $\widehat{\mathcal{M}}$, Demonstrations $\mathcal{D}_{\text{demo}}$
    \State Initialize replay buffer $\mathcal{D}_{\text{cap}}$ and $\text{Dict}_{\text{visit}}$ (Eq.~\eqref{DictDefine})
    % \State Explorer Policy $\pi^E \leftarrow$ Behaviour clone on $\mathcal{D}_{\text{demo}}$
    % \State Goal predictor $\mathcal{P}_\phi$ used for evaluation $\leftarrow$ Trained on $\mathcal{D}_{\text{demo}}$(Eq.~\eqref{eq:train_goal_predictor})
    \State Train explorer policy $\pi^E$ using Behavior Cloning on $\mathcal{D}_{\text{demo}}$
    \State Train goal predictor $\mathcal{P}_\phi$ on $\mathcal{D}_{\text{demo}}$ (Eq.~\eqref{eq:train_goal_predictor})
    % \State Init$\text{Dict}_{\text{visit}} \leftarrow \{seed : [0]^{|demo|} \mid$ for $seed$ of every trajectory $\mathcal{D}_{\text{demo}}[seed] \}$
    \For{$n = 1$ to $N_{train}$}
        \State Initialize empty trajectory $\tau$
        \State Randomly sample a demonstration $\tau^{(i)} \in \mathcal{D}_{\text{demo}}$
        \State Initialize the environment to the initial state of $\tau^{(i)}$ %$s$ = env.reset(seed=$seed_i$)
        % \State $Demo_{use}$ = $\mathcal{D}_{\text{demo}}[seed]$
        \State Sample a capability-aware goal $g$ by Algorithm \ref{Algorithm-FindGoal}
        \For{$t = 0$ to $L_{\tau}$}\label{alg:CollectTrajectory}
            \If{agent has not reached $g$ and $t < T_{go}$}
                \State $\pi = \pi^G(s, g)$
            \Else
                \State $\pi = \pi^E(s)$
            \EndIf
            \State Step in the real environment using $\pi$ and add this step to $\tau$
            % \State $\text{Dict}_{\text{visit}}[seed][i] += 1$ for every $obs_i$ in $Demo_{use}$ if If\_reach($obs$, $obs_i$)
            \State Update $\text{Dict}_{\text{visit}}[\tau^{(i)}]$ (Eq.~\eqref{DictUpdate})

        \EndFor
        \State $\mathcal{D}_{\text{cap}} \leftarrow \mathcal{D}_{\text{cap}} \cup \{\tau\}$
        \State Update $\widehat{\mathcal{M}}$ with $\mathcal{D}_{\text{cap}}$ \label{alg:learnM}
        \State Update $\pi^G$ using imagined rollouts with $\widehat{\mathcal{M}}$ \label{alg:learnG}
    \EndFor
\end{algorithmic}
\end{algorithm}
\end{minipage}
}
\end{wrapfigure}

\textbf{World Model and Policy Training.} \tool{} stores the trajectories generated under the Go-Explore paradigm with capability-aware goal sampling in a dataset $\mathcal{D}_{\text{cap}} = \{(s_t,a_t,s_{t+1})_{t=1}^{T}\}$ for world model and policy training. A predictive world model $\widehat{\mathcal{M}}$ approximates the transition dynamics \(\mathcal{T}(s'|s, a)\) in the real world \(\mathcal{M}\) as \(\widehat{\mathcal{T}}(s'|s, a)\). 
%These rollouts in $\mathcal{D}_{\text{cap}}$ are inherently aligned with the agent’s current capability, ensuring that the resulting trajectories are of higher quality and more feasible than those targeting final goals beyond the agent's reach. 
Our model learning algorithm is based on the Dreamer backbone~\citep{hafner2019dream,hafner2020mastering,hafner2023mastering}, which updates the world model $\widehat{\mathcal{M}}$ via supervised learning using $\mathcal{D}_{\text{cap}}$.
% \begin{equation}
% \widehat{\mathcal{M}} \leftarrow \arg\min_{\theta} \mathbb{E}_{(s,a,s') \sim \mathcal{D}_{\text{cap}}} \left\| \widehat{\mathcal{M}}_\theta(s, a) - s' \right\|^2,
% \end{equation}
Once the world model is updated, we train the goal-conditioned policy $\pi^G$ using imagined trajectories generated by the world model $\widehat{\mathcal{M}}$. 
Intuitively, since $\mathcal{D}_{\text{cap}}$ is collected by exploring around demonstration states in $\mathcal{D}_{\text{demo}}$, the learned model enables the agent to generate imagined trajectories that remain grounded in task-relevant regions of the state space.
Each imagined trajectory from the learned world model $\widehat{\mathcal{M}}$ begins at $s_0$, a state randomly sampled from a trajectory $\tau$ in $\mathcal{D}_{\text{demo}} \cup \mathcal{D}_{\text{cap}}$, and is rolled out for $H$ steps using the goal-conditioned policy $\pi^G(a_t | s_t, g)$. The goal state $g$ is selected as a future state $s_H$ from the same trajectory $\tau$. %, i.e., $g$ lies $H$ steps ahead of $s_0$ in the real environment. 
The objective is to train $\pi^G$ to reinforce trajectories that efficiently reach $g$ in the imagined rollouts from 
$s_0$ under the learned dynamics $\widehat{\mathcal{M}}$. To achieve this, we adopt an actor-critic algorithm that leverages a self-supervised temporal distance function $D_t(s, g)$~\citep{mendonca2021discovering}, which estimates the number of steps required to transition from state $s$ to goal $g$. The reward function is defined as: $r^G(s, g) = -D_t(s, g)$.
% \begin{equation}\label{eq: reward_function}
% r^G(s, g) = -D_t(s, g)
% \end{equation}
This formulation encourages the policy to generate actions that minimize the estimated temporal distance to the goal.
The temporal distance estimator \(D_t\) is trained by extracting state pairs \((s_t, s_{t+k})\) from simulated trajectories generated by the world model. The function learns to predict the normalized temporal difference between two states: $D_t\big(\Psi(s_t), \Psi(s_{t+k})\big) \approx \frac{k}{H}$, 
% \begin{equation}\label{eq: temporal_distance}
% D_t\big(\Psi(s_t), \Psi(s_{t+k})\big) \approx \frac{k}{H}
% \end{equation}
where \(\Psi\) denotes a transformation applied to states (e.g., embedding them into the world model's latent space), and \(H\) is the length of the generated rollout. More details on the model-based learning algorithm and the full training procedure for \(D_t\) can be found in Appendix~\ref{subs: rssm} and Appendix~\ref{subs: Dt-training}.

\begin{wrapfigure}{r}{0.35\textwidth}
\vspace{-15pt}
  \centering
    \includegraphics[width=0.35\textwidth]{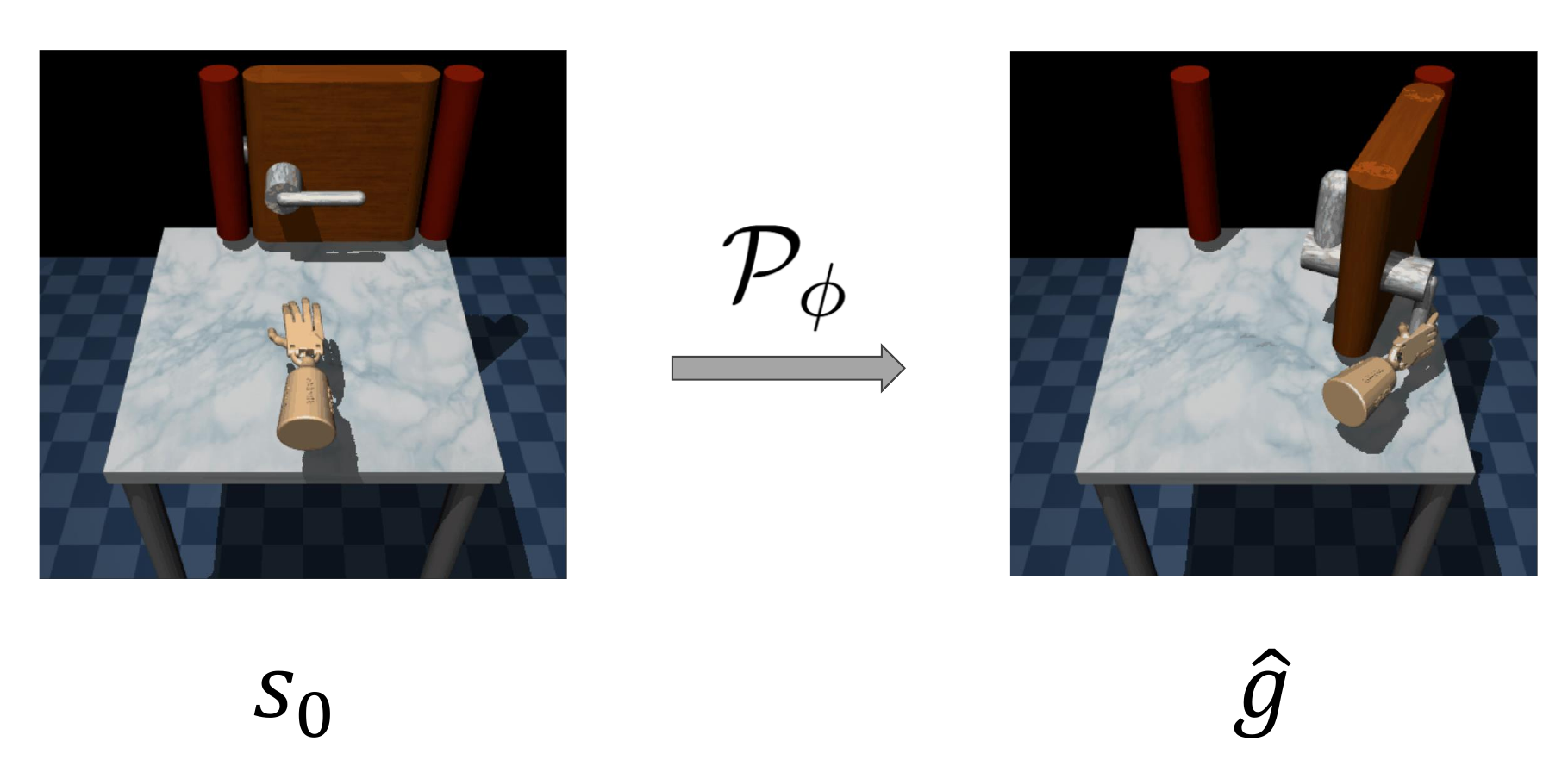}
    % \vspace{-0.5cm}
    % \subfigure[]{\includegraphics[width=0.25\textwidth]{E.png}}
    % \subfigure[]{\includegraphics[width=0.25\textwidth]{r.png}}
  \caption{The workflow of the goal predictor $\mathcal{P}_\phi$.}
  %\vspace{-50pt}
  \label{fig:goal-predictor}
\end{wrapfigure}

\textbf{Goal Predictor.} %In our setting, although each observation inherently contains goal-related information (e.g., target object position), %As a result, we define the goal space to be the entire observation space, rather than a manually specified subspace. This formulation captures not only the target's location but also other essential contextual features, such as the robot arm's velocity, joint positions, and environmental states.
At training time, the goal-conditioned policy $\pi^G(\cdot \mid \cdot, g)$ is trained using intermediate states from demonstration trajectories as goal conditions (recall $\mathcal{S} = \mathcal{G}$). This assumes access to demonstrations, with the final states used as the target goal condition. However, at test time, this assumption no longer holds: for unseen scenarios, the true final goal state is not available. This raises the challenge of how to specify an appropriate goal condition based solely on the agent's current observation. We introduce a goal predictor $\mathcal{P}_\phi$, a learned model that infers a goal state $\hat{g}$ given the current observation $s$:
\begin{equation}
\mathcal{P}_\phi : s \mapsto \hat{g}, \quad \text{where } \hat{g} = \mathcal{P}_\phi(s)
\end{equation}
The mapping learned by $\mathcal{P}_\phi$ is illustrated in Figure~\ref{fig:goal-predictor}. 
It is trained using demonstration trajectories $\mathcal{D}_{\text{demo}}$, by minimizing the mean squared error between the predicted goal and the true final observation:
\begin{equation}
\label{eq:train_goal_predictor}
\min_\phi \ \mathbb{E}_{(\tau^{(i)} = s_0^{(i)},\ldots, s_{L}^{(i)}) \sim \mathcal{D}_{\text{demo}}} \left\| \mathcal{P}_\phi(s_t^{(i)}) - s_{L}^{(i)} \right\|_2^2
\end{equation}
Once trained, the goal predictor enables $\pi^G$ to generalize to new tasks. Given a test-time state $s^{\text{test}}$, the predicted goal $\hat{g}^{\text{test}} = \mathcal{P}_\phi(s^{\text{test}})$ serves as the planning target for the agent $\pi = \pi^G(\cdot \mid s^{\text{test}}, \hat{g}^{\text{test}})$. The complete training pipeline of \tool{} is detailed in Algorithm~\ref{Algorithm-TrainingFrame}.

\textbf{Rationale Behind \tool{}'s Design.} Let $\mathcal{J}(\pi,\mathcal{M})$ and $\mathcal{J}(\pi^{e}, \mathcal{M})$ be the expected return of the agent's policy $\pi$ and expert policy $\pi^{e}$ in the real-world MDP $\mathcal{M}$. We want to bound their return difference:
\begin{equation}
    \label{eq:objective1}
    \min_\pi \ \ \left| \mathcal{J}(\pi^{e}, \mathcal{M}) - \mathcal{J}(\pi, \mathcal{M}) \right|,
\end{equation}
Let $R_{\text{max}}$ be the maximum of the reward with unknown dynamics: $R_{\text{max}} = \max_{(s,a)}\mathcal{R}(s,a)$ and $\rho^{\pi}_{\mathcal{M}}(s, a) = (1-\gamma) \sum_{t=0}^{\infty} \gamma^t P(s_t = s, a_t = a)$ be the discounted state-action visitation distribution of a policy $\pi$ in the real world MDP $\mathcal{M}$. Suppose that the total variation of learned dynamics model $\widehat{\mathcal{M}}$ from the true transitions $\mathcal{M}$ is bounded such $\mathbb{D}_{\text{TV}}(\mathcal{T}(s,a), \widehat{\mathcal{T}}(s,a)) \leq \alpha \quad \forall (s,a) \in \mathcal{S}\times\mathcal{A}$. According to previous work \citep{rafailov2021visual, demoss2023ditto, kolev2024efficient}, we have:%we can bound the gap of expected return in real world $\mathcal{M}$ between policy $\pi$ and expert policy $e$ as:
\begin{equation}\label{returnbound}
    \left| \mathcal{J}(\pi^{e}, \mathcal{M}) - \mathcal{J}(\pi, \mathcal{M}) \right| \leq 
    \underbrace{\alpha \frac{R_{\text{max}}}{(1-\gamma)^2}}_{\text{model prediction error}} +
    \underbrace{\frac{R_{\text{max}}}{1-\gamma} \mathbb{D}_{\text{TV}}\left( \rho^{\pi^{e}}_\mathcal{M}, \rho^\pi_{\widehat{\mathcal{M}}}\right)}_{\text{adaptation error}}
\end{equation}
where $\rho^{\pi^{e}}_\mathcal{M}$ is the discounted visitation distribution of the expert policy and $\mathbb{D}_{\text{TV}}$ denote total variation distance. %This bound indicates that the performance of the sub-optimal learned policy is bounded with a bias that is proportional to the model learning error tied to how accurately the model captures the true dynamics. 
The model prediction error with respect to the true environment dynamics can be reduced by collecting more real-world data. In contrast, the adaptation error depends on the total variation distance between the distribution of trajectories generated by policy $\pi$ under the learned world model $\widehat{\mathcal{M}}$ and the expert distribution under the true dynamics $\mathcal{M}$. 
% From the triangle inequality of the total variation (TV) distance, we have
% \begin{equation}\label{eq:cargoobj}
% \mathbb{D}_{\text{TV}}(\rho^{\pi^{e}}_\mathcal{M}, \rho^\pi_{\widehat{\mathcal{M}}}) 
% \leq 
% \mathbb{D}_{\text{TV}}(\rho^{\pi^{e}}_\mathcal{M}, \rho^{e}_{\widehat{\mathcal{M}}}) 
% + 
% \mathbb{D}_{\text{TV}}(\rho^{\pi^{e}}_{\widehat{\mathcal{M}}}, \rho^\pi_{\widehat{\mathcal{M}}}).  
% \end{equation}
%Thus, to reduce the total variation distance between the visitation distribution of the learned policy $\pi$ under the learned world model $\widehat{\mathcal{M}}$, and that of the expert in the real-world MDP $\mathcal{M}$, 
Thus, the learning problem reduces to bounding the deviation between the behavior of the learned policy $\pi$ under the learned model and the expert behavior under the true environment.
%we can reduce the learning problem to bound the deviation of the learned policy $\pi$'s behavior in the learned model from the the expert behavior in the true environment ($\rho^{\pi^{e}}_{{\mathcal{M}}}$ vs $\rho^\pi_{\widehat{\mathcal{M}}}$).
To this end, given any $H$-step trajectory $(s_0, s_1, \ldots, s_H)$ sampled from expert demonstrations $\mathcal{D}_{\text{demo}}$, \tool{} encourages the agent to match expert behavior by rewarding it for reaching the final state $g = s_H$ starting from $s_0$ under the learned dynamics model $\widehat{\mathcal{M}}$ (Line~\ref{alg:learnG} in Algorithm~\ref{Algorithm-TrainingFrame}).
%\tool{} rewards the agent for matching the expert behavior over $\widehat{\mathcal{M}}$ at . 

% Thus, we optimize the objective as:
% \begin{equation}
%     \label{eq:objective2}
%     \min_\pi \ \ \mathbb{D}_{\text{TV}}(\rho^\pi_{\widehat{\mathcal{M}}}, \rho^e_\mathcal{M}),
% \end{equation}

%However, directly matching the full expert visitation distribution can be challenging, especially when the agent’s current policy is far from optimal. 

\newtheorem{theorem}{Theorem}

We further show that \tool{} effectively reduces the model prediction error by leveraging the BC explore policy $\pi^E = \pi^\text{BC}$ for data collection. In the following, we use $d_t^{\mathcal{M},\pi}$ to denote the marginal state-action distribution at time $t$ induced by policy $\pi$ in the environment $\mathcal{M}$. We assume $d_t^{\mathcal{D}_{\text{demo}}} \approx d_t^{\mathcal{M}, \pi^{e}}$, where $\mathcal{D}_{\text{demo}}$ is the dataset of demonstrations generated by the expert $\pi^{e}$, and is sufficiently representative to approximate the true marginal distributions at each timestep. Assuming: (1) $\pi^{\text{BC}}$ accurately approximates the expert policy in $\mathcal{D}_{\text{demo}}$, (2) the world model $\widehat{\mathcal{M}}$ is accurately trained on state transitions induced by $\pi^{\text{BC}}$, and (3) the learned policy $\pi$ generates trajectories in $\widehat{\mathcal{M}}$ that closely match the expert's behavior, we can bound the model prediction error along the imagined rollouts generated by $\pi$ under $\widehat{\mathcal{M}}$:
\begin{theorem}
\label{theorem}
Let \( \mathcal{M} \) denote the true dynamics model and \( \widehat{\mathcal{M}} \) the learned model. Let \( \pi_{\mathrm{BC}} \) be a behavior-cloned policy, and \( \pi \) a new policy. Let \( \mathcal{D}_{\text{demo}} \) be a dataset of expert demonstrations from an unknown expert policy. Suppose that, for all \( t = 0, 1, \dots, T \), (1) Closeness of behavior cloning: $\mathbb{D}_{\emph{TV}}\left( d_t^{\mathcal{M}, \pi_{\mathrm{BC}}}, d_t^{\mathcal{D}_{\text{demo}}} \right) \leq \kappa$, (2) Model learning error under BC: $\E_{(s,a) \sim d_t^{\mathcal{M}, \pi_{\mathrm{BC}}}} \left[ \mathbb{D}_{\emph{TV}}\left( \mathcal{M}(\cdot \mid s,a), \widehat{\mathcal{M}}(\cdot \mid s,a) \right) \right] \leq \mu$, and (3) Trajectory distribution closeness: $\mathbb{D}_{\emph{TV}}\left( \rho^{\pi^{e}}_\mathcal{M}, \rho^\pi_{\widehat{\mathcal{M}}} \right) \leq \nu$.
Then for all \( t = 0, 1, \dots, T \), we have:
\[
\E_{(s,a) \sim d_t^{\widehat{\mathcal{M}}, \pi}} \left[ \mathbb{D}_{\emph{TV}}\left( \mathcal{M}(\cdot \mid s,a), \widehat{\mathcal{M}}(\cdot \mid s,a) \right) \right] \leq \mu + 2\kappa + 2\nu.
\]
\end{theorem}

\section{Experiments}
\label{Sec:Experiments}

% We evaluate the \tool{} on various robotic manipulation environments, aiming to answer the following three questions: (1) Does \tool{} outperform other baselines using demonstrations in other ways? (2) Dose \tool{} realize capability-aware goal sampling gradually, achieving the agent learning progress? (3) Is BC-Explorer necessary in \tool{}?

We evaluate \tool{} across a diverse set of challenging robotic manipulation environments to address the following research questions:
(Q1) Does \tool{} outperform existing imitation learning baselines that leverage demonstrations in alternative ways?
(Q2) Can \tool{} effectively realize capability-aware goal sampling that aligns with the agent's learning progress?
(Q3) How essential are the proposed capability-aware goal sampling and BC-Explorer components to the overall performance of \tool{}?

%\subsection{Environments}

\textbf{Environments}. For our experiments, we evaluate and compare \tool{} against several baselines across three robot environment suites with sparse rewards: MetaWorld ~\citep{yu2020meta}, Adroit~\citep{rajeswaran2017learning}, and Maniskill~\citep{gu2023maniskill2, taomaniskill3}. We adopt the five "very hard" level environments from MetaWorld, as categorized by~\citet{seo2023masked}: Shelf Place, Disassemble, Stick Pull, Stick Push, Pick Place Wall. These environments are considered the most challenging tasks in Metaworld, requiring precise robotic arm control with only sparse task completion rewards. We also use three dexterous hand manipulation tasks from the Adroit suite: Door, Hammer, Pen. To succeed in these three environments, the agent must perform fine-grained and intricate finger manipulations, enabling the grasping and movement of different objects. We also selected three challenging tasks from the ManiSkill benchmark: PegInsertionSide, StackCube, and PullCubeTool. The sparse-reward ManiSkill environments are the most difficult tasks in our benchmarks due to their high-dimensional state and action spaces. During training, we used only 10 demonstration trajectories per task for the MetaWorld and Adroit environments, and 20 demonstration trajectories per task for the ManiSkill environments. More details about each task can be found in Appendix~\ref{Sec:Environment Detail}.

\textbf{Baselines}. Our approach is developed on top of the \textbf{Dreamer} framework~\citep{hafner2019dream,hafner2020mastering,hu2023planning,duan2024exploring, duan2024learning}, making it a key model-based RL baseline for evaluating the performance gains of \tool{}. %Dreamer learns a latent world model and trains a goal-conditioned policy using an actor-critic algorithm. It samples trajectories by setting the final observation of a demonstration as the goal, while resetting the environment using the same seed as the corresponding demonstration. 
% Jump-Start Reinforcement Learning (\textbf{JSRL})~\citep{uchendu2023jump} is a curriculum-based approach that leverages a pre-trained expert policy derived from offline data to guide early-stage sampling. At each training episode, it initially follows the expert policy for a number of steps determined by curriculum progress, after which the online policy takes over. While JSRL aims to improve the quality of online data, its performance strongly depends on the availability and quality of offline expert data, which may be limited in real-world settings. 
Jump-Start Reinforcement Learning (\textbf{JSRL})~\citep{uchendu2023jump} is a curriculum-based approach that leverages a guide-policy pretrained from offline data to guide early-stage exploration during online training. At the beginning of each training episode, the agent follows the guide-policy for a number of steps determined by curriculum progression, after which control is handed over to the online policy. %While JSRL also aims to enhance the quality of online data collection, its effectiveness is highly contingent on the availability and quality of the offline expert data—resources that are often scarce in real-world scenarios.
In our JSRL implementation, due to the limited number of demonstrations available, training a reliable guide-policy becomes challenging. Therefore, we directly use the demonstration trajectories as the guide-policy. Specifically, we reset the environment to a demonstration initial state, enabling the agent to replicate expert behavior during the initial phase of each episode before switching to the online policy. Our JSRL implementation is also built on top of the Dreamer framework.
\textbf{MoDem}~\citep{hansen2022modem} represents one of the most efficient frameworks in the model-based RL literature. It pretrains its policy using a small set of demonstrations and repeatedly oversamples the demonstrations to train both the world model and the policy. We consider MoDem to be the strongest baseline due to its fast convergence and low data requirements. \textbf{Cal-QL}~\citep{nakamoto2023cal} is a state-of-the-art algorithm following the offline-to-online RL paradigm. It uses demonstrations to pretrain the $Q$-function and applies calibration to mitigate performance drop when transitioning from offline to online learning phases. In addition to these representative baselines, we compare against four more imitation learning baselines: \textbf{GAIL}~\citep{ho2016generative}, \textbf{PWIL}~\citep{dadashi2020primal}, \textbf{SQIL}~\citep{reddy2019sqil}, \textbf{ValueDice}~\citep{kostrikov2019imitation}, and \textbf{RLPD}~\citep{ball2023efficient}, a well-tuned variant of SAC that leverages offline data when learning online, in the Appendix~\ref{subsec:more-baselines}.

\begin{figure}[t] 
  \centering
    \subfigure[Disassemble]{\includegraphics[width=0.24\textwidth]{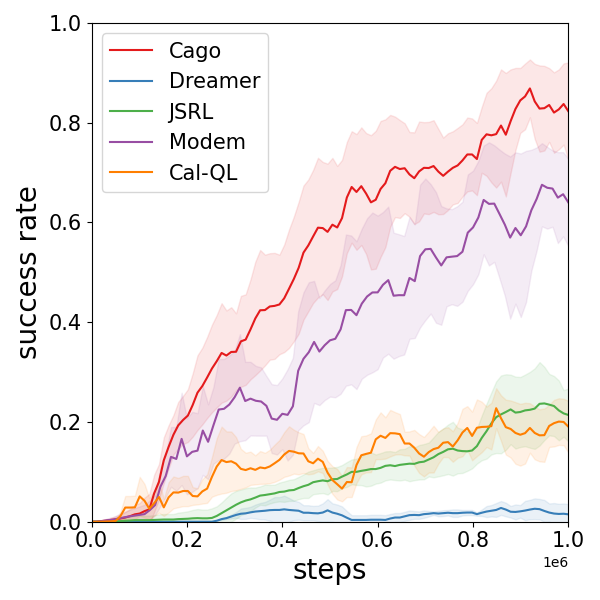}}
    \subfigure[PickPlaceWall]{\includegraphics[width=0.24\textwidth]{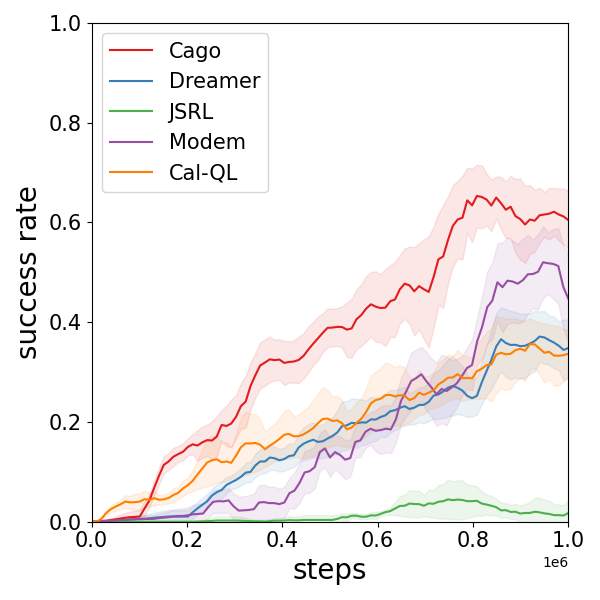}}
    \subfigure[ShelfPlace]{\includegraphics[width=0.24\textwidth]{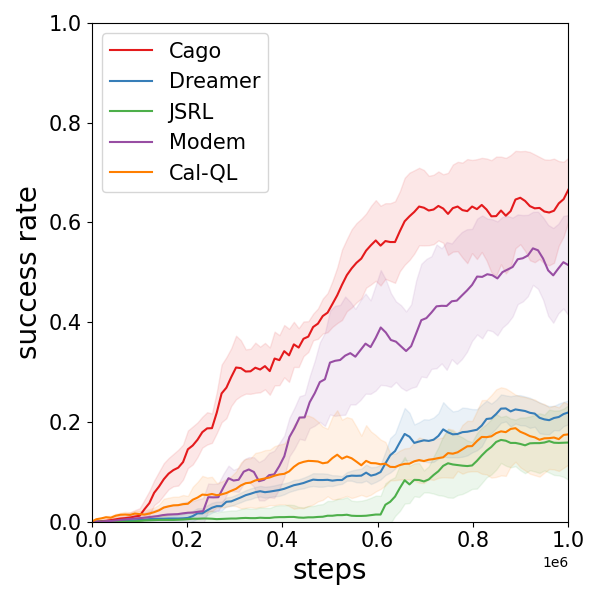}}
    \subfigure[StickPull]{\includegraphics[width=0.24\textwidth]{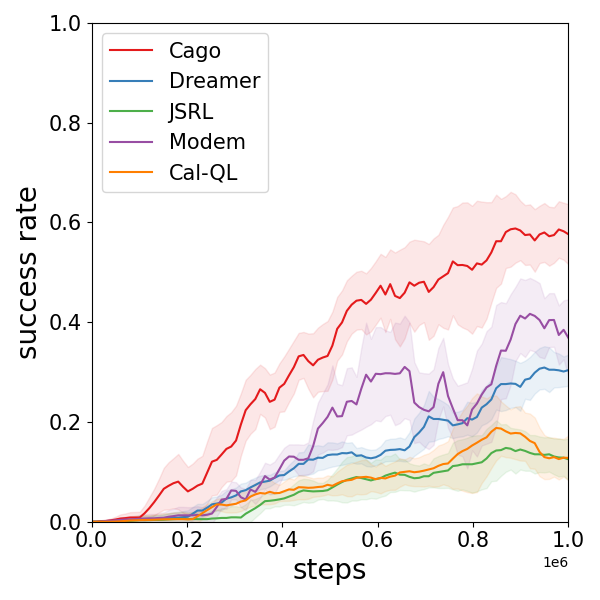}}
    \subfigure[StickPush]{\includegraphics[width=0.24\textwidth]{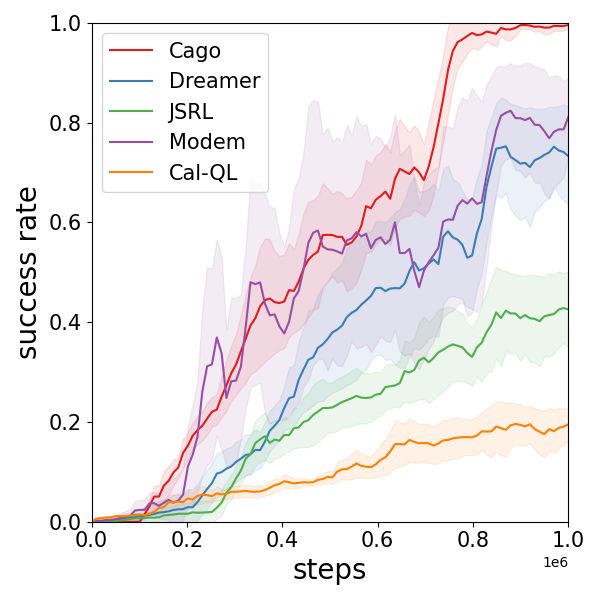}}
    \subfigure[Door]{\includegraphics[width=0.24\textwidth]{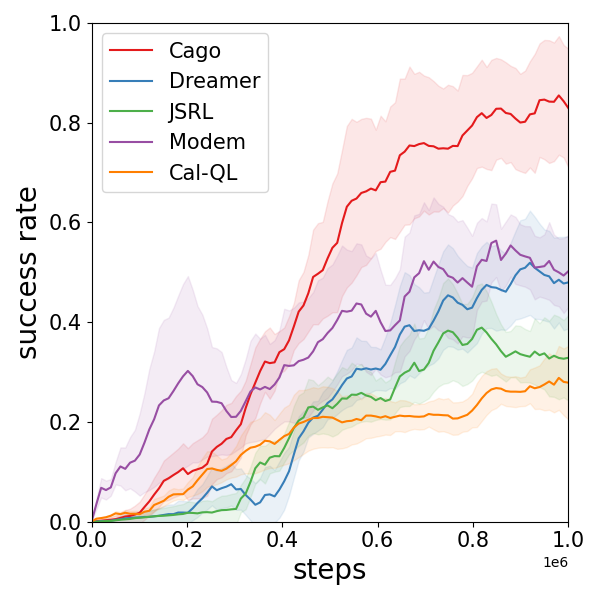}}
    \subfigure[Hammer]{\includegraphics[width=0.24\textwidth]{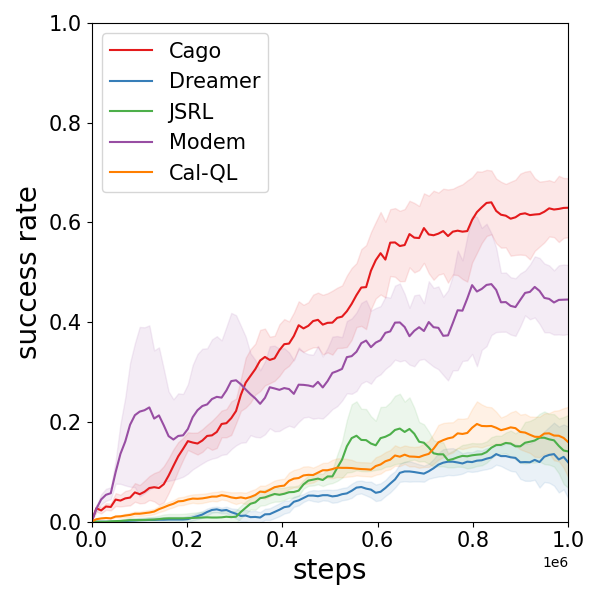}}
    \subfigure[Pen]{\includegraphics[width=0.24\textwidth]{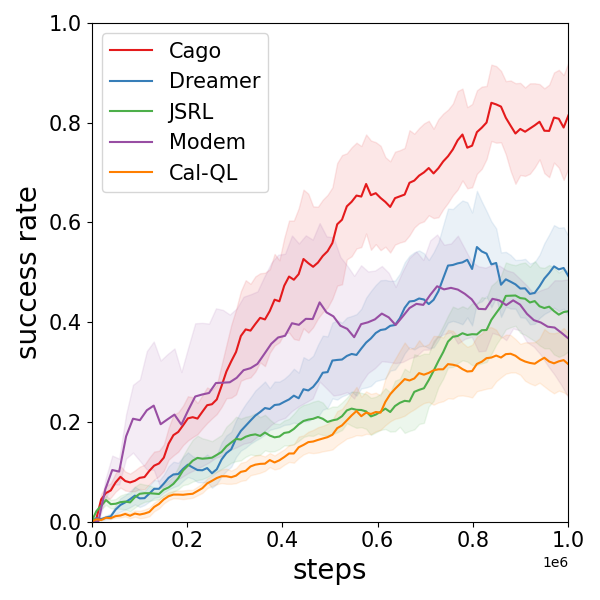}}
    \subfigure[PullCubeTool]{\includegraphics[width=0.24\textwidth]{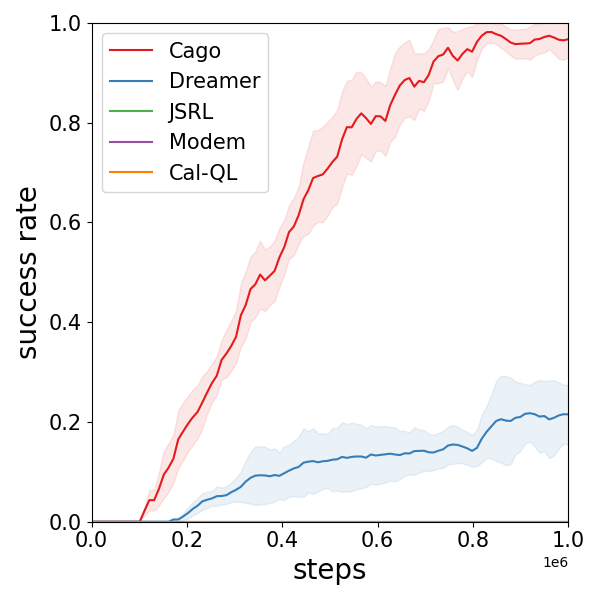}}
    \subfigure[PegInsertionSide]{\includegraphics[width=0.24\textwidth]{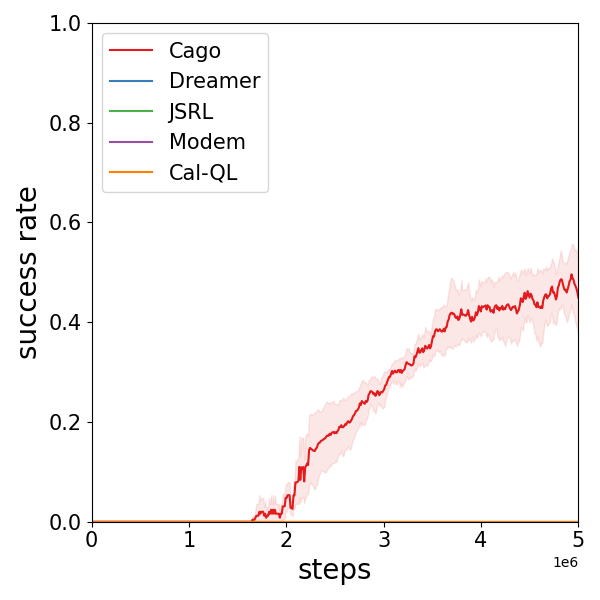}}
    \subfigure[StackCube]{\includegraphics[width=0.24\textwidth]{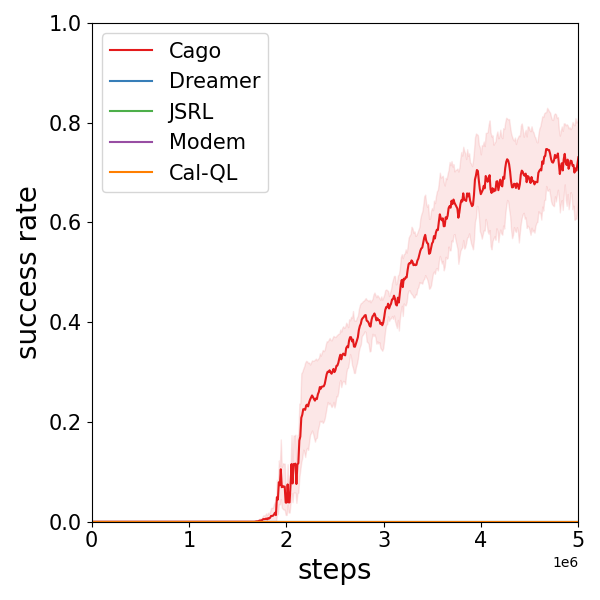}}
  \caption{Experiment results comparing \tool{} with the baselines over 8 random seeds. The solid line denotes the average success rate in \emph{evaluation}, while the shaded region signifies the standard deviation.}
  \vspace{-10pt}
  \label{fig:exp-results}
\end{figure}

% During training, we reset the environment using the same seeds as in the demonstrations to ensure identical initial conditions. This allows us to use the final observation from each demonstration as the goal during evaluation. However, to evaluate the generalization ability of \tool{}, we must test it on unseen seeds. For these unseen seeds, the final goal is unknown, as the goal space is equivalent to the observation space in our setting. Nonetheless, the initial observation after resetting the environment contains some goal-relevant information. Therefore, we train a goal predictor using only demonstration data to infer the final goal from the initial observation. This predictor enables goal inference for arbitrary seeds during evaluation. We evaluate all methods on 100 random environment seeds that were not included in the demonstrations and calculate the average success rate for each evaluation round.

%we reset the environment using the same seeds as in the demonstrations to ensure identical initial conditions. This allows us to use the final observation from each demonstration as the goal during evaluation. However, 

\textbf{Main Results}. During training, we uniformly sample a demonstration and reset the environment using the same seed that was used to collect it. All baselines share the same seeds and demonstration data. To evaluate generalization, we test on unseen environment seeds. For these, \tool{} uses the goal predictor (Section~\ref{subsec: framework}) to infer goal conditions for the goal-conditioned policy $\pi^G$. Each method is evaluated on 100 held-out seeds, and we report the average success rate over these episodes.
Figure~\ref{fig:exp-results} depicts the mean learning performance of \tool{} and all baselines in terms of the agent's task success rate averaged over 8 random seeds. On the MetaWorld very hard tasks, \tool{} consistently outperforms all the baselines in both final performance and learning efficiency. In the Adroit suite, although Modem exhibits rapid early learning due to its behavior cloning (BC) pretraining and oversampling strategy, \tool{} surpasses it in final performance after 1e6 environment interaction steps. Notably, Dreamer, which shares the same world model and policy architecture as \tool{}, performs significantly worse, underscoring the effectiveness of the capability-aware goal sampling strategy. Our JSRL baseline, based on the same world model architecture, adopts a uniform curriculum to reduce the guide-steps from demonstrations. It lags behind \tool{} in both learning speed and final success rate, highlighting the effectiveness of our goal-sampling strategy in adapting to the agent's evolving capabilities. In the ManiSkill environments, given the limited demonstrations, \tool{} stands out as the only method capable of achieving high success rates.

\begin{figure}[t] 
  \centering
    \subfigure[Disassemble]{\includegraphics[width=0.24\textwidth]{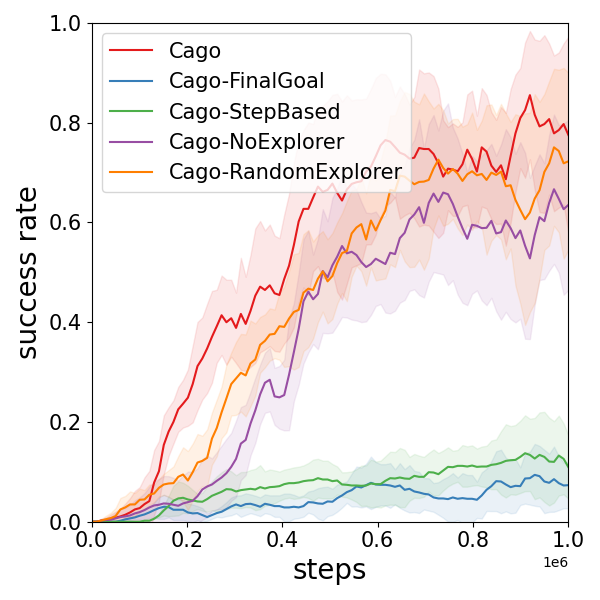}}
    \subfigure[StickPush]{\includegraphics[width=0.24\textwidth]{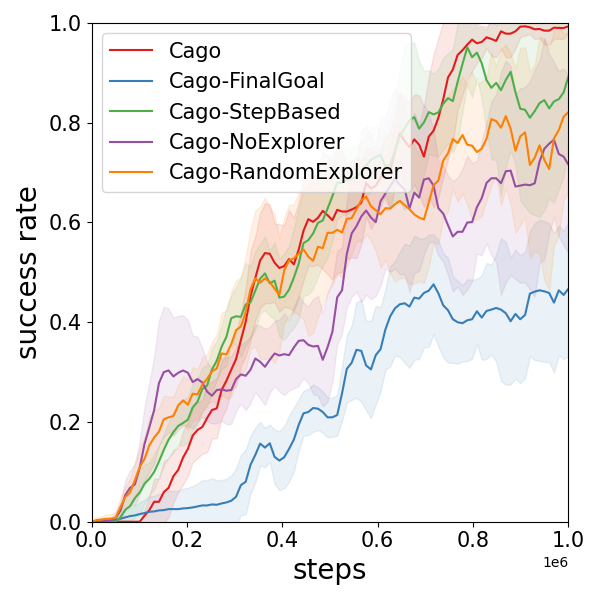}}
    \subfigure[Pen]{\includegraphics[width=0.24\textwidth]{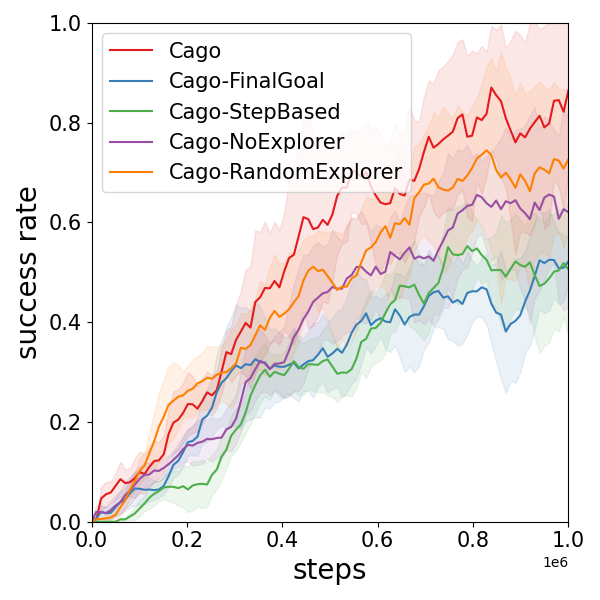}}
    \subfigure[Sampling Progress]{\label{fig:goal-distribution}\includegraphics[width=0.24\textwidth]{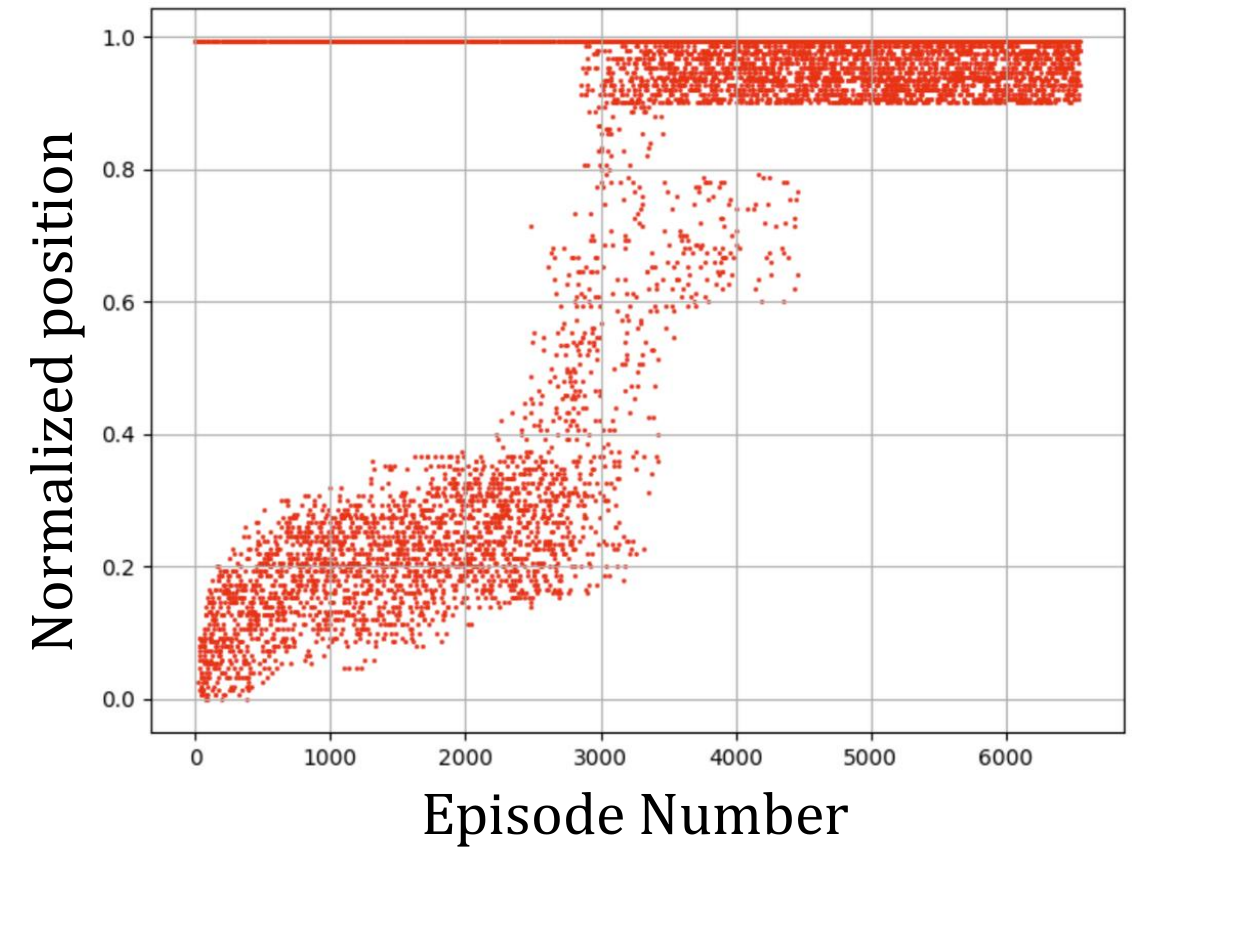}}
  \caption{Figure(a),(b),(c) are the results of ablation study on the importance of each component of \tool{} over 5 seeds. Figure(d) shows the progress of capability-aware goal sampling in Stickpush.}
  \vspace{-10pt}
  \label{fig:ablation-results}
\end{figure}

\textbf{Capability-Aware Goal Distribution}. To answer Q2, we visualize the progression of capability-aware goal sampling throughout the training process in the StickPush environment in Figure~\ref{fig:goal-distribution}. Each red dot represents the normalized position of a sampled goal within a demonstration trajectory, with 0 indicating the start and 1.0 indicating the final demonstration state. Early in training, the agent predominantly samples goals at lower normalized positions, focusing on subgoals near the beginning of the trajectory that are within its current capabilities. As training advances, goal sampling gradually shifts toward higher normalized positions, indicating the agent's increasing ability to pursue more challenging goals closer to task completion.
%This adaptive sampling strategy enables \tool{} to track the agent’s skill level and adjust goal difficulty accordingly. 
By continuously targeting goals just at the boundary of the agent's current capability, \tool{} facilitates efficient learning in sparse-reward, long-horizon tasks. %where naive or uniform goal sampling would often fail to provide meaningful training signals.

% In the ablation experiments, we are going to evaluate how essential the BC-Explorer component is to the overall performance of \tool{}. Because we follow the Go-Explore sampling paradigm, the trajectories' quality should rely on both the go-phase and explore-phase. In the go-phase, we apply capability-aware goal sampling. And the explore-phase uses the BC-Explorer. How important is the BC-Explorer, can we get similar performance without it? We conduct an ablation version of \tool{} called \tool{}-NoExplorer, which only has the go-phase. We conduct the ablation experiment on three environments: ...

%In our ablation study, we aim to investigate the individual contributions of (a) capability-aware goal sampling and (b) the BC-Explorer component to the overall performance of \tool{}. The first ablation \tool{}-FinalGoal only performs part (b) for BC Exploration. During exploration, it does not consider the agent's capability of goal reaching and always selects the final observation of the demonstration for which the environment is initialized to. The BC explorer $\pi^E$ takes over $\pi^G$ at half of the rollout.  Our second ablation \tool{}-NoExplorer only performs part (a) for capability-aware goal sampling and furthur explore the environment using the BC Explorer upon reaching a sampled goal. The last ablation \tool{}-RandomExplorer replaces the BC Explorer with a policy that samples actions uniformly at random from the action space in the Explore phase of our Go-Explore sampling paradigm. 
\textbf{Ablation Studies}. To answer Q3, we assess the individual contributions of (a) capability-aware goal sampling and (b) the BC-Explorer component to the overall performance. The first ablation, \tool{}-FinalGoal, retains only (b): it uses BC-based exploration but always selects the final observation from a demonstration in the goal phase of our Go-Explore sampling paradigm, ignoring the agent’s current goal-reaching capability. The BC Explorer takes control from the goal-conditioned policy halfway through each rollout. The second ablation, \tool{}-StepBased, also retains (b), but samples goals from demonstrations in proportion to training steps (current training step / predefined total training steps). However, it does not assess the agent's actual capabilities, and may therefore sample goals that are either too easy or too difficult for the agent at its current learning stage. The third ablation, \tool{}-NoExplorer, keeps only (a): it uses capability-aware goal sampling, but does not explore beyond the sampled goal with the BC Explorer. The fourth ablation, \tool{}-RandomExplorer, replaces the BC Explorer with a uniformly random policy during the exploration phase of our Go-Explore-style rollout strategy. We conduct the ablation study on the Disassemble and StickPush tasks from MetaWorld, and the Pen task from Adroit. As shown in Figure~\ref{fig:ablation-results}, removing capability-aware goal sampling significantly degrades performance. Without it, the agent often enters the Explore phase from states far outside the demonstration region, making it difficult for the BC-Explorer to make meaningful progress. The BC-Explorer itself is also crucial, as it accelerates learning by generating high-quality exploratory rollouts. We further examine how the number and quality of demonstrations (including suboptimal ones), as well as the hyperparameter $\lambda_{visit}$ and $\delta$, affect performance; see Appendix~\ref{sec:more-experiments} for details.

\begin{figure}[H] 
  \centering
    \subfigure[Door]{\includegraphics[width=0.3\textwidth]{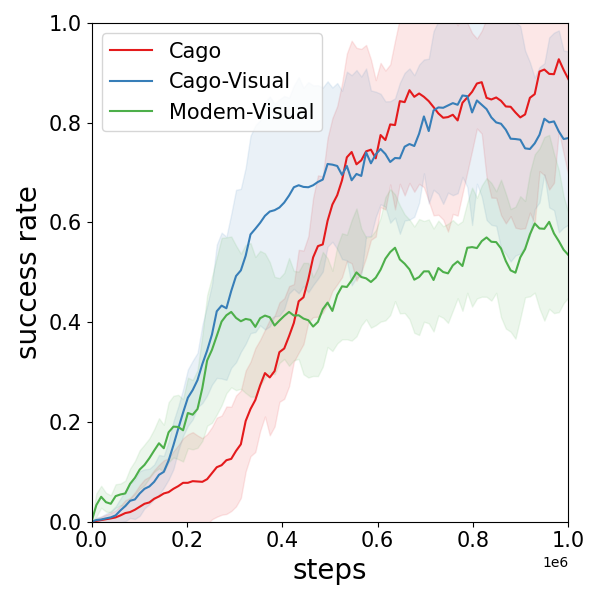}}
    \subfigure[Hammer]{\includegraphics[width=0.3\textwidth]{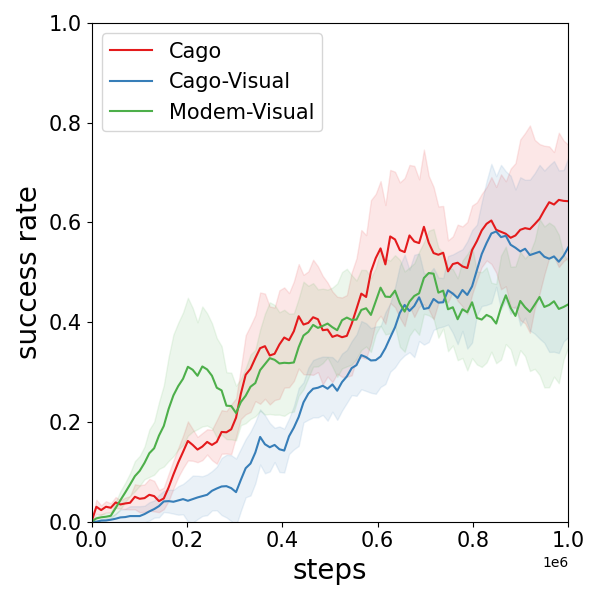}}
    \subfigure[Pen]{\includegraphics[width=0.3\textwidth]{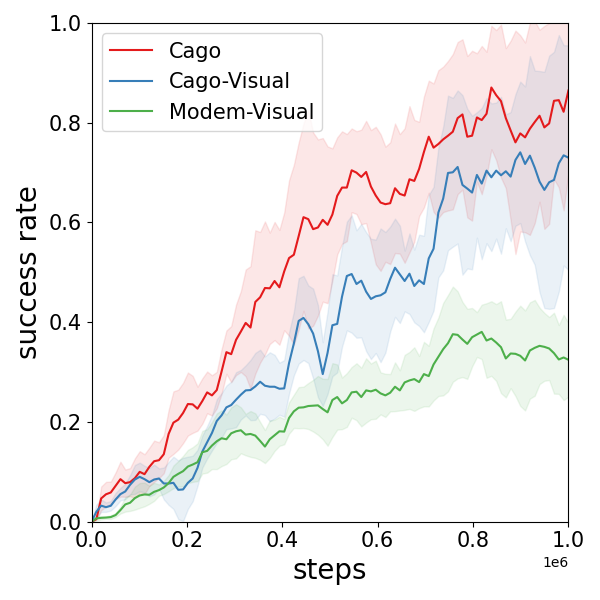}}
  \caption{Visual input experiment results over 5 random seeds.}
  % \vspace{-10pt}
  \label{fig:visual-results}
\end{figure}

\textbf{Visual-input Environments}. We further assess \tool{}'s applicability in high-dimensional visual settings. Specifically, we extend our framework to raw pixel observations by replacing the vector-based states and goals with RGB image inputs of size (64, 64, 3), resulting in a variant referred to as \tool{}-Visual. In this setting, both the policy and the goal predictor $\mathcal{P}_\phi$ operate on image representations. We benchmark \tool{}-Visual against Modem-Visual, a strong model based baseline that similarly utilizes image-based observations and demonstration. As shown in Figure~\ref{fig:visual-results}, \tool{}-Visual not only retains performance similar to the original \tool{}, but also consistently outperforms Modem-Visual, highlighting the robustness of our method in visual domains. Details of the goal predictor employed in our visual-input experiments are provided in Appendix~\ref{subsec: visual-gp}.

\section{Related Work}
\label{Sec:RelatedWork}

%Demonstrations have played a central role in improving the efficiency and stability of reinforcement learning (RL), with prior works leveraging them in diverse ways across key components of the RL pipeline~\citep{arulkumaran2024pragmatic, nair2018overcoming}. A prominent line of work utilizes demonstrations for direct learning, typically through Behavior Cloning (BC) or its variants~\citep{bain1995framework, torabi2018behavioral}.
\textbf{Imitation Learning.} Demonstrations are a key tool for improving the efficiency of RL, with prior work integrating them across various stages of the RL pipeline~\citep{arulkumaran2024pragmatic, nair2018overcoming, cui2025abstraction}. A prominent approach uses demonstrations for direct learning via Behavior Cloning (BC) and its variants~\citep{bain1995framework, torabi2018behavioral}.
%Jump-Start Reinforcement Learning (JSRL)~\citep{uchendu2023jump} leverages a pre-trained expert policy derived from offline data to guide early-stage sampling. 
%MoDem~\citep{hansen2022modem} utilizes only a small number of demonstrations to pretrain its policy and repeatedly oversamples these demonstrations to train both the world model and the policy.
The introduction of the Generative Adversarial Imitation Learning (GAIL) algorithm~\citep{ho2016generative} has driven significant advances in scalable deep imitation learning methods~\citep{fu2017learning,ghasemipour2020divergence,kostrikov2018discriminator,jena2020,FinnCAL16,BlondeK19,OrsiniRHVDGGBPA21,eysenbach2021replacing}. Beyond adversarial approaches, several imitation learning algorithms aim to match the state action distributions of the expert and the agent through non-adversarial techniques, such as non-parametric models~\citep{KimP18}, random network distillation~\citep{WangCAD19}, support estimation~\citep{BrantleySH20}, Wasserstein distance minimization~\citep{dadashi2020primal}, and moment matching~\citep{SwamyCBW21}. Demonstrations have also been the bridge between imitation learning and offline RL. Conservative Q-learning (CQL)~\citep{kumar2020conservative} and Cal-QL~\citep{nakamoto2023cal} regularize $Q$-values using demonstration data to better estimate out-of-distribution actions. Traditional imitation learning typically requires direct learning from demonstrations, either by recovering a reward function from demonstrations, or matching expert and agent state-action distributions (e.g. GAIL\citep{ho2016generative}), or incorporating demonstrations as anchors to mitigate overestimation and instability caused by out-of-distribution actions (e.g. Cal-QL\citep{nakamoto2023cal}). In contrast, \tool{} introduces a novel use of demonstrations by treating them as a structured roadmap for building an adaptive curriculum, scaffolding the agent's learning to enable steady progress toward solving the full task. 
%Another major use of demonstrations is in reward learning, particularly through inverse reinforcement learning (IRL). Representative works, including Generative Adversarial Imitation Learning ~\citep{ho2016generative}, PWIL~\citep{dadashi2020primal} and AdRIL~\citep{eysenbach2021replacing} aim to infer reward signals that explain expert behavior. 

\textbf{Curriculum for Learning from Demonstrations.} 
Several prior works have explored curriculum design using demonstration. \citet{yengera2021curriculum,TaoSC024} introduce difficulty scores to rank demonstrations, offering a theoretical framework for selecting optimal trajectories to scaffold learning. Task Phasing~\citep{bajaj2023task} automatically extracts curriculum phases from demonstrations and dynamically transitions the agent through them during training. %Reverse Forward Curriculum Learning~\citep{TaoSC024} proposes a bidirectional strategy that adaptively refines task sampling and ordering to improve demonstration and sample efficiency with state resetting. 
JSRL~\citep{uchendu2023jump} design a curriculum-based approach that leverages a guide-policy pretrained from demonstration to guide early-stage exploration during online training. Compared to these curriculum learning methods using demonstration, \tool{} employs a \emph{goal-level} curriculum that incrementally samples intermediate goals from demonstrations based on the agent's evolving capabilities. 
% In many real-world settings, agents face sparse reward environments—settings where rewards are only provided upon successful task completion—making exploration especially difficult~\citep{liu2022goal, plappert2018multi, ghosh2019learning}. To address this inherent exploration difficulties in sparse-reward reinforcement learning environments,
A number of methods have explored the use of expert demonstration states as informative priors for guiding agent learning. A common strategy involves resetting the agent from states sampled along demonstration trajectories~\citep{nair2018overcoming,PengALP18,HosuR16}, enabling the agent to experience these expert regions of the state space without exhaustive exploration. Some approaches employ structured curricula to sequence these resets, either through manually designed progressions~\citep{Zhu0MRECTKHFH18} or automated strategies such as reverse curricula, which gradually increase the exploration horizon by training the agent from goal states backward~\citep{Resnick2018,Salimans2018,TaoSC024}. While effective in simulation, these methods rely on the ability to precisely reset the environment to arbitrary demonstration states—a requirement that poses significant challenges in physical systems, where replicating exact configurations, including latent dynamics like joint velocities, is often infeasible. Rather than forcibly placing the agent into expert states—a strong form of intervention—\tool{} encourages the agent to actively reach intermediate goals that are sampled to match its current competence. This self-directed learning process allows the agent to internalize problem-solving skills more effectively, promoting deeper understanding through its own attempts rather than directly resetting. While demonstrations provide useful guidance, true mastery requires the agent to explore and overcome challenges through trial and error.

% For a broader discussion of related work on goal-conditioned RL and state resets, see Appendix~\ref{sec:extend-related-work}.

%Curriculum learning has emerged as a powerful strategy for improving sample efficiency and stability in reinforcement learning by structuring the learning process through gradually increasing task difficulty. 
%Early works like C-Planning~\citep{zhang2021c} propose automatic curricula for goal-reaching tasks by an automatic generation of intermediate waypoints to reach distant goals. 
%demonstrations to inform curriculum design. 
%Jump-Start Reinforcement Learning (JSRL)~\citep{uchendu2023jump} also follows a curriculum perspective, gradually shifting control from a pre-trained expert policy to the online policy. 

%This fine-grained guidance enables more effective exploration and experience collection during training. 

\section{Conclusion}

We introduce \tool{}, a novel method that leverages demonstrations in a dynamic, goal-guided manner to tackle exploration challenges in sparse-reward environments. By continuously monitoring the agent’s capabilities, \tool{} constructs an adaptive curriculum that incrementally samples intermediate goals from demonstrations, effectively scaffolding learning and enabling steady progress toward solving the full task.
Extensive experiments show consistent improvements over baselines. A key limitation of \tool{} lies in its reliance on resetting environments to the initial states specified in the demonstrations. Appendix~\ref{sec:discussion} provides a discussion on how this requirement could be relaxed to improve applicability in real-world settings.

%A key limitation of \tool{} is its reliance on high-quality demonstrations. In future work, we aim to extend \tool{} to handle suboptimal offline data, enabling robust learning in more practical scenarios.

%This targeted guidance not only enhances exploration efficiency but also facilitates the learning of long-horizon, complex tasks that are otherwise difficult to solve with conventional RL techniques. Our method demonstrates that demonstrations, when used beyond static initialization or reward inference, can serve as a powerful tool for structuring training and improving data quality throughout the learning process. 
 %We believe this work opens new directions for incorporating limited demonstration data into the reinforcement learning pipeline in more strategic and adaptive ways.

\section*{Reproducibility Statement} 

% The code for \tool{} is available on \url{}. For hyperparameter settings, please refer to Appendix~\ref{Sec:hyperparameters}.

The code for \tool{} is available at \href{https://github.com/RU-Automated-Reasoning-Group/Cago}{\texttt{https://github.com/RU-Automated-Reasoning-Group/Cago}}. For hyperparameter settings, please refer to Appendix~\ref{Sec:hyperparameters}.

\section*{Acknowledgements}

We thank the anonymous reviewers for their comments and suggestions. This work was supported by NSF Award \#CCF-2124155.

% \section*{Acknowledgements}

% We thank the anonymous reviewers for their comments and suggestions.

\bibliography{Reference/Reference}
\bibliographystyle{apalike}

%%%%%%%%%%%%%%%%%%%%%%%%%%%%%%%%%%%%%%%%%%%%%%%%%%%%%%%%%%%%
% Start the appendix

\newpage 
\appendix
\addcontentsline{toc}{section}{Appendix} 
{\fontsize{20}{14}\selectfont \textbf{Appendix}}
\parttoc

\section{Proof of Theorem 1}
\label{sec:proof}

\textbf{Theorem 1}.
Let \( \mathcal{M} \) denote the true dynamics model and \( \widehat{\mathcal{M}} \) the learned model. Let \( \pi_{\mathrm{BC}} \) be a behavior-cloned policy, and \( \pi \) a new policy. Let \( \mathcal{D}_{\text{demo}} \) be a dataset of expert demonstrations from an unknown expert policy. Suppose that, for all \( t = 0, 1, \dots, T \), (1) Closeness of behavior cloning: $\mathbb{D}_{\emph{TV}}\left( d_t^{\mathcal{M}, \pi_{\mathrm{BC}}}, d_t^{\mathcal{D}_{\text{demo}}} \right) \leq \kappa$, (2) Model learning error under BC: $\E_{(s,a) \sim d_t^{\mathcal{M}, \pi_{\mathrm{BC}}}} \left[ \mathbb{D}_{\emph{TV}}\left( \mathcal{M}(\cdot \mid s,a), \widehat{\mathcal{M}}(\cdot \mid s,a) \right) \right] \leq \mu$, and (3) Trajectory distribution closeness: $\mathbb{D}_{\emph{TV}}\left( \rho^{\pi^{e}}_\mathcal{M}, \rho^\pi_{\widehat{\mathcal{M}}} \right) \leq \nu$.
Then for all \( t = 0, 1, \dots, T \), we have:
\[
\E_{(s,a) \sim d_t^{\widehat{\mathcal{M}}, \pi}} \left[ \mathbb{D}_{\emph{TV}}\left( \mathcal{M}(\cdot \mid s,a), \widehat{\mathcal{M}}(\cdot \mid s,a) \right) \right] \leq \mu + 2\kappa + 2\nu.
\]

\begin{proof}
By triangle inequality on expectations and total variation:
\begin{align*}
&\E_{(s,a) \sim d_t^{\widehat{\mathcal{M}}, \pi}} \left[ \TV\left( \mathcal{M}(\cdot \mid s,a), \widehat{\mathcal{M}}(\cdot \mid s,a) \right) \right] \\
&\leq \E_{(s,a) \sim d_t^{\mathcal{D}_{\text{demo}}}} \left[ \TV\left( \mathcal{M}(\cdot \mid s,a), \widehat{\mathcal{M}}(\cdot \mid s,a) \right) \right] + 2 \, \TV\left( d_t^{\widehat{\mathcal{M}}, \pi}, d_t^{\mathcal{D}_{\text{demo}}} \right).
\end{align*}

We now bound each term separately:

\begin{itemize}
\item For the first term, apply triangle inequality again:
\begin{align*}
\E_{(s,a) \sim d_t^{\mathcal{D}_{\text{demo}}}} \left[ \TV\left( \mathcal{M}(\cdot \mid s,a), \widehat{\mathcal{M}}(\cdot \mid s,a) \right) \right]
&\leq \E_{(s,a) \sim d_t^{\mathcal{M}, \pi_{\mathrm{BC}}}} \left[ \TV\left( \mathcal{M}(\cdot \mid s,a), \widehat{\mathcal{M}}(\cdot \mid s,a) \right) \right] \\
&\quad + 2\, \TV\left( d_t^{\mathcal{D}_{\text{demo}}}, d_t^{\mathcal{M}, \pi_{\mathrm{BC}}} \right) \\
&\leq \mu + 2\kappa,
\end{align*}
using assumptions (1) and (2).

\item For the second term, use that marginal total variation is bounded by trajectory total variation:
\[
\TV\left( d_t^{\widehat{\mathcal{M}}, \pi}, d_t^{\mathcal{D}_{\text{demo}}} \right) \leq \TV\left( \rho^{\widehat{\mathcal{M}}, \pi}, \rho^{\mathcal{M},\pi^{e}} \right) \leq \nu,
\]
by assumption (3).
\end{itemize}

Combining:
\[
\E_{(s,a) \sim d_t^{\widehat{\mathcal{M}}, \pi}} \left[ \TV\left( \mathcal{M}(\cdot \mid s,a), \widehat{\mathcal{M}}(\cdot \mid s,a) \right) \right] \leq \mu + 2\kappa + 2\nu.
\]
\end{proof}

\section{Extended Background}

\subsection{Dreamer World Model}
\label{subs: rssm}
The RSSM consists of an encoder, a recurrent model, a representation model, a transition predictor, and a decoder, as formulated in Equation~\ref{eq:world_model}.
And it employs an end-to-end training methodology, where its parameters are jointly optimized based on the loss functions of various components, 
including dynamic transition prediction, reward prediction, and observation encoding-decoding. 
These components often operate in a latent space rather than the original observation space, as encoded by the World Model. 
Therefore, during end-to-end training, the losses of all components indirectly optimize the latent space. 

The encoder $f_E$ encodes the input state $x_t$ into a embed state $e_t$, which is then fed with the deterministic state $h_t$ into the representation model $q_\varphi$ to generate the posterior state $z_t$.
The transition predictor $p_\varphi$ predicts the prior state $\hat{z}_t$ based on the deterministic state $h_t$ without access to the current input state $x_t$.
Using the concatenation of either $(h_{t}, z_{t})$ or $(h_{t}, \hat{z}_t)$ as input, the recurrent transition function $f{\varphi}$ iteratively updates the deterministic state $h_t$ with given action $a_t$.

\begin{equation}\label{eq:world_model}
  \begin{aligned}
    \text{Encoder:\qquad} & e_t = f_E(e_t | x_t)\\
    \text{Recurrent model:\qquad} & h_t = f_\varphi(h_{t-1}, z_{t-1}, a_{t-1}) \\
    \text{Representation model:\qquad} & z_t \sim q_\varphi(z_t | h_t, e_t) \\
    \text{Transition predictor:\qquad} & \hat{z}_t \sim p_\varphi(\hat{z}_t | h_t) \\
    \text{Decoder:\qquad} & \hat{x}_t \sim f_D(\hat{x}_t | h_t, z_t)
  \end{aligned}
\end{equation}

\subsection{Temporal Distance Training in LEXA}
\label{subs: Dt-training}

The goal-reaching reward $r^G$ is defined by the self-supervised temporal distance objective~\citep{mendonca2021discovering} which aims to minimize the number of action steps needed to transition from the current state to a goal state within imagined rollouts.
We use $b_t$ to denote the concatenate of the deterministic state $h_t$ and the posterior state $z_t$ at time step $t$.

\begin{equation}\label{eq: temporal_distance_reward1}
  b_t = (h_t, z_t)
\end{equation}

The temporal distance $D_t$ is trained by sampling pairs of imagined states $b_t, b_{t+k}$ from imagined rollouts and predicting the action steps number $k$ between the embedding of them, 
with a predicted embedding $\hat{e}_t$ from $b_t$ to approximate the true embedding $e_t$ of the observation $x_t$.

\begin{equation}\label{eq: temporal_distance_reward2}
  \text{Predicted embedding:\qquad} emb(b_t) = \hat{e}_t \approx e_t, \qquad \text{where\quad} e_t = f_E(x_t)
\end{equation}

\begin{equation}\label{eq: temporal_distance_reward3}
  \text{Temporal distance: } D_t(\hat{e}_t, \hat{e}_{t+k}) \approx k/H \qquad \text{where\quad} \hat{e}_t = emb(b_t)\quad \hat{e}_{t+k} = emb(b_{t+k})
\end{equation}

\begin{equation}\label{eq: temporal_distance_reward4}
  r^G_t(b_t, b_{t+k}) = -D_t(\hat{e}_t, \hat{e}_{t+k})
\end{equation}

\section{Discussion}
\label{sec:discussion}

In this section, we discuss several key questions and extensions regarding the design, applicability, and limitations of \tool{}. Our goal is to clarify how \tool{} fundamentally differs from existing imitation learning paradigms, what design choices drive its effectiveness, and how it can be adapted or extended to broader settings. We also explain and clarify its current assumptions—such as initial state resetting and model-based training. Through this discussion, we aim to provide deeper insights into the generality, scalability, and future potential of \tool{} as a new framework for learning from demonstrations.

\textbf{1. How does \tool{} differ from traditional Imitation Learning methods?}

\tool{} is fundamentally different from traditional Imitation Learning (IL) methods, both in methodology and motivation. IL typically requires direct learning from demonstrations, either by recovering a reward function from demonstrations, or matching expert and agent state-action distributions (e.g. GAIL\citep{ho2016generative}), or incorporating demonstrations as anchors to mitigate overestimation and instability caused by out-of-distribution actions (e.g. Cal-QL\citep{nakamoto2023cal}). In contrast, \tool{} presents a new paradigm to utilize demonstrations: it dynamically tracks the agent's competence along demonstrated trajectories and uses this signal to select intermediate states in the demonstrations that are just beyond the agent's current reach as goals to guide online trajectories collecting. To evaluate this novel perspective, we compared \tool{} against state-of-the-art baselines that represent diverse strategies for learning from demonstrations, including distribution mathcing (e.g. GAIL\citep{ho2016generative} and PWIL\citep{dadashi2020primal}), curriculum learning (e.g. JSRL\citep{uchendu2023jump}), model-based exploration (e.g. MoDem\citep{hansen2022modem}), and offline-to-online fine-tuning (e.g. CalQL\citep{nakamoto2023cal} and RLPD\citep{ball2023efficient}). %These comparisons are essential to highlight the effectiveness of \tool{} as a new paradigm for learning from demonstrations.

\textbf{2. Can \tool{} be extended with more sophisticated similarity metrics to handle complex tasks?}

In our implementation of \tool{}, we use the observation space as the goal space and adopt simple similarity metrics such as MSE or L2 distance to demonstrate the applicability and robustness of \tool{}'s curriculum strategy in both state-based and pixel-based environments. \tool{} can be naturally extended to support more complex settings. For example, in high-dimensional visual environments, \tool{} could adopt more expressive similarity metrics such as SSIM (Structural Similarity Index) or FSIM (Feature Similarity Index). In tasks involving semantic or language-conditioned goals, one could compute a scalar score function that estimates how well a visual state satisfies a goal, and then compare states based on the difference in their scores with respect to the same goal. This score function can be instantiated using pretrained vision-language models, allowing goal representations to move beyond raw states to language embeddings or other semantic forms. Our main contribution is a general framework for goal sampling from demonstrations—based on the agent's capability—which can be paired with any goal space and similarity metric appropriate to the domain. 

\textbf{3. How does \tool{} perform relative to curriculum learning baselines?}

In our experiments, we included Jump-Start Reinforcement Learning (JSRL\citep{uchendu2023jump}), a recent and strong curriculum learning baseline. JSRL pretrains a guide policy on offline data to provide a curriculum of starting states for the RL policy, which significantly simplifies the exploration problem. As the RL policy improves, the effect of the guide-policy is gradually diminished, leading to an RL-only policy that is capable of further autonomous improvement. We discuss four additional curriculum learning baselines: \cite{yengera2021curriculum, bajaj2023task, dai2021automatic, hermann2020adaptive}.
\cite{yengera2021curriculum} constructs a personalized demonstration curriculum by computing task-specific difficulty scores based on the teacher’s and learner’s policies. In contrast, \tool{} does not require hand-crafted difficulty metrics—instead, it extracts intermediate goals directly from the demonstration, and guides the agent to reach them progressively based on its own learning progress. This avoids the challenge of designing difficulty metrics for complex manipulation tasks, which is non-trivial without expert knowledge. \cite{bajaj2023task} assumes access to a demonstrator policy, either via imitation learning or inverse reinforcement learning. However, our experiments show that standard imitation learning performs poorly on our benchmarks, limiting the effectiveness of such approaches in our domain. \cite{dai2021automatic, hermann2020adaptive} segment demonstrations and construct a curriculum by resetting episodes to intermediate states along the demonstration trajectory, starting from the end and moving backward as learning progresses. \tool{} explicitly avoids such state resets, which are often unrealistic in real-world robotic settings without simulator support. we conducted additional experiments comparing \tool{} to \textbf{ACED}\citep{dai2021automatic} and \textbf{Task phasing}\citep{bajaj2023task} on the Fetch-PickPlace and Fetch-Slide environments that they were evaluated on. The tables~\ref{tab:cago_curriculum_camp} below report the final success rates after 1M training steps, averaged over 5 random seeds (We were unable to run ACED on Fetch-Slide). 

\begin{table}[H]
\centering
\caption{Comparison of \tool{} and ACED, Task phasing across Fetch-PickPlace and Fetch-Slide tasks.}
\label{tab:cago_curriculum_camp}
\begin{tabular}{lccc}
\toprule
\textbf{Task} & \textbf{Cago (Ours)} & \textbf{ACED}~\citep{dai2021automatic} & \textbf{Task phasing}~\citep{bajaj2023task} \\
\midrule
Fetch-PickPlace & 1.00 & 0.93 & 0.75 \\
Fetch-Slide     & 0.48 & --   & 0.20 \\
\bottomrule
\end{tabular}
\end{table}

\textbf{4. Is the need for environment resets in \tool{} a fundamental limitation?}

Indeed, resetting to specific initial states is a limitation of \tool{}, but as we mentioned in the Section~\ref{Sec:RelatedWork}, it is already a substantial improvement over arbitrary state resets, which are often infeasible in real-world robotics. Importantly, \tool{} does not require exact resets to the full demonstration initial states. For instance, in the Peg Insertion task, the peg's initial position can differ from the demonstration, and \tool{} can learn to reach states within the early portion of the demonstration. However, some elements—like the hole and the box—are non-movable and randomly initialized by the environment. Since the agent cannot manipulate these objects, we have to reset them to match the demonstration configuration to ensure goal reachability. The constraint over other controllable components, such as the robot arm and peg, can be relaxed. This partial reset strategy reduces the burden and makes \tool{} more practical for real-world settings.

\textbf{5. What motivates the choice of a model-based RL formulation for \tool{}, rather than a model-free alternative?}

In early experiments, we attempted to train \tool{} in a model-free manner by learning the goal-conditioned policy directly from real environment rollouts. But we further find that instead of directly using $\mathcal{D}_{\text{cap}}$ to train $\pi^G$ in Model-free manner, the world model can offer a richer set of imagined rollouts to train the $\pi^G$, which improves the learning efficiency. In our setting, the simulated trajectories have starting states and goal states randomly sampled from the same trajectory in $\mathcal{D}_{\text{cap}}$. As a result, the simulated trajectories still resemble segments from $\mathcal{D}_{\text{cap}}$, while significantly increasing trajectory diversity and data richness. The MBRL—leveraging simulated trajectories allows us to generate more diverse and abundant training data. This promotes generalization across the environment. \tool{} builds upon the \textbf{Dreamer} framework~\citep{hafner2019dream,hafner2020mastering,hu2023planning,duan2024exploring,duan2024learning}, which serves as the model-based cornerstone of our approach. Dreamer learns a latent world model and trains a goal-conditioned policy using an actor-critic algorithm. It samples trajectories by setting the final observation of a demonstration as the goal, while resetting the environment using the same seed as the corresponding demonstration. \textbf{PEG}~\citep{hu2023planning} introduces an alternative goal-selection strategy that enhances exploration within the Dreamer framework. To demonstrate the improvement brought by \tool{} over its cornerstone, we conduct comparisons against both Dreamer and PEG. Table~\ref{tab: PEG_comparison} presents the performance comparison of \tool{} against Dreamer and PEG across three environments(results are averaged over 5 random seeds). \tool{} clearly surpasses its model-based cornerstone, Dreamer, as well as the competitive goal-picking strategy, PEG, across all environments.

\begin{table}[H]
\centering
\caption{Performance comparison of \tool{} against Dreamer and PEG.}
\label{tab: PEG_comparison}
\begin{tabular}{lccc}
\toprule
Environment & \tool{} (Ours) & Dreamer & PEG \\
\midrule
StickPush    & 0.99 & 0.68 & 0.76 \\
Disassemble   & 0.80 & 0.02 & 0.04 \\
Adroit-Pen    & 0.82 & 0.52 & 0.54 \\
\bottomrule
\end{tabular}
\end{table}

\textbf{6. Can \tool{} be applied to broader scenarios—such as real-world settings with diverse initial states and uncertain dynamics?}

\tool{} can be extended to support diverse initial states and uncertain dynamics.
For diverse initial states in real-world settings, \tool{} can gradually prioritize those where the agent has the most learning potential—starting from demonstration initial states and expanding to nearby ones where the agent can still benefit from the demonstration trajectories for dynamic capability tracking and progressive goal selection. As training progresses, we incorporate successful rollouts from these nearby initial states into the demonstration set, broadening its coverage. This enables a natural curriculum that eventually transitions to diverse, real-world initial states.
For uncertain, stochastic dynamics, \tool{} remains robust by not requiring exact reproduction of actions in demonstration trajectories. Instead, it adapts goals based on observed agent successes, making the curriculum self-correcting in the face of noise or variability. If the agent consistently fails to reach a specific demonstration state, \tool{}'s capability metric naturally shifts goal sampling toward more achievable targets. Additionally, goal selection can be extended with probabilistic estimates—such as goal-reaching likelihood or future state predication uncertainty from an ensemble of world models—which we view as a promising direction for high-stochasticity domains.

\textbf{7. Does \tool{} risk overfitting to a limited number of demonstrations?}

\tool{} mitigates overfitting to a small number of demonstrations by training its goal-conditioned policy $\pi^G$ within a world model. Rather than training $\pi^G$ to reach states solely from demonstrations, \tool{} instructs $\pi^G$ to learn to reach diverse states sampled from real rollouts in $\mathcal{D}_{\text{cap}}$, which records the trajectories generated under our Go-Explore paradigm with capability-aware goal sampling, in its world model. This promotes generalization across the environment. To evaluate the generalization capability, we tested \tool{} on 500 unseen initial states generated from random seeds, each differing from those in the demonstrations. We report both the average L2 deviation from demonstration initial states and the average success rate across 500 unseen seeds across
three environments in Table~\ref{tab:init_deviation_success}. The results are consistent with Figure~\ref{fig:exp-results}.

\begin{table}[H]
\centering
\caption{Initial demo observation deviation and average success rate across 500 unseen seeds.}
\label{tab:init_deviation_success}
\begin{tabular}{lcc}
\toprule
Environment & Average Initial Observation Deviation & Average Success Rate(500 seeds) \\
\midrule
StickPush   & 0.0547 & 0.982 \\
Disassemble  & 0.1405 & 0.806 \\
Adroit-Pen   & 1.2749 & 0.824 \\
\bottomrule
\end{tabular}
\end{table}

\section{Limitations and Future Work}
\label{sec:limitation}

While \tool{} shows promising results in leveraging demonstrations to improve exploration and learning efficiency in sparse-reward, long-horizon tasks, several limitations remain. In our experiments, we employ image MSE similarity(or L2 distance for state) and a simple counting scheme, which are generalizable and applicable across diverse environments, as illustrative measures of evaluating agent capability in \tool{}. However, in much more complex environments, such simple approaches may not constitute the optimal strategy. In the future, incorporating uncertainty-aware models for both agent capability estimation and subgoal selection could improve robustness and adaptability of \tool{} framework. Extending \tool{} to multi-agent or real-world robotics scenarios is another promising direction, where the complexity of coordination and physical constraints introduces new challenges for efficient demonstration utilization and goal guidance.

\section{Environment Details}
\label{Sec:Environment Detail}

We evaluate and compare \tool{} against several baselines across three robot environment suites using sparse rewards: MetaWorld~\citep{yu2020meta}, Adroit~\citep{rajeswaran2017learning}, and Maniskill~\citep{gu2023maniskill2, taomaniskill3}. In this section, we provide more details about each benchmark and the specific experimental setup.

\subsection{MetaWorld}

\begin{figure}[h]
    \centering
    \includegraphics[width=1\textwidth]{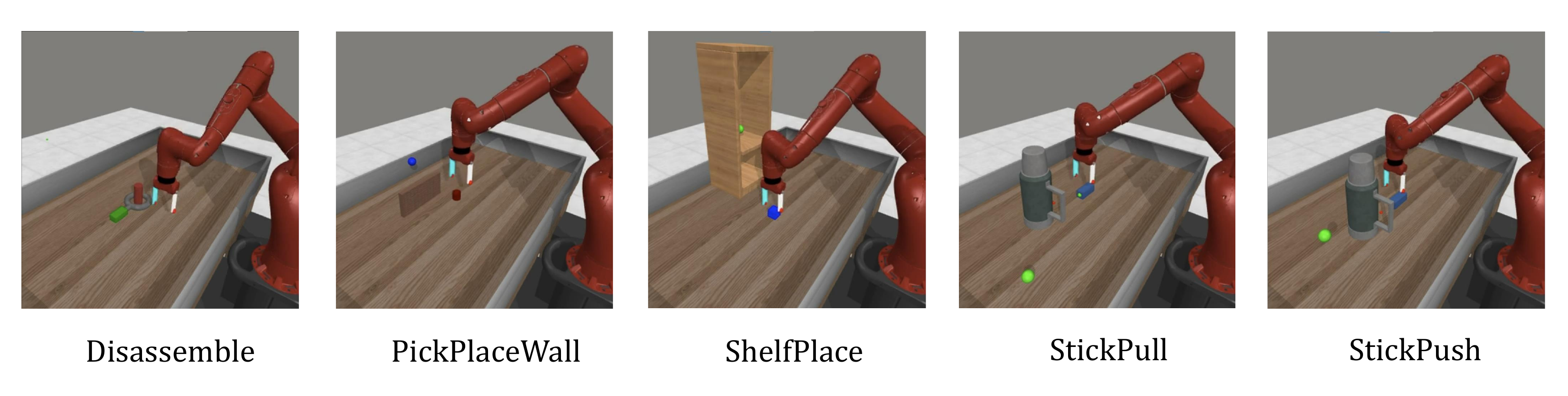}
    \caption{5 very hard level environments from MetaWorld}
    \label{fig:Env-MetaWorld}
\end{figure}

MetaWorld~\citep{yu2020meta} is a widely used benchmark suite designed to evaluate the generalization and manipulation capabilities of reinforcement learning algorithms in robotic control tasks. It consists of a diverse collection of continuous control environments simulated using the MuJoCo physics engine. Each task requires a robotic arm to interact with objects in the scene to achieve goal-directed behavior under sparse or dense reward settings. MetaWorld includes 50 distinct tasks of varying difficulty, ranging from simple reaching to complex object manipulation. In our experiments, we focus on the "very hard" subset of tasks identified in prior studies~\citep{seo2023masked, hansen2022modem}, which are characterized by sparse rewards, delayed feedback, and the need for precise low-level control, making them particularly suitable for benchmarking sample efficiency and generalization in demonstration-augmented learning frameworks. The five very hard tasks we choose are: Shelf Place, Disassemble, StickPull, Stick Push, Pick Place Wall. Shelf Place: The agent must grasp an object and place it accurately onto a shelf, requiring precise vertical and lateral arm control. Disassemble: The task involves picking a nut out of the a peg, demanding a strong grasp and directional pulling motion. Stick Pull: The robot needs to grasp a stick and pull a bottle, requiring fine force control and coordinated motion. Stick Push: The goal is to grasp a stick and push a bottle, emphasizing controlled contact and alignment with the target location. Pick Place Wall: The agent must pick up an object and place it over a wall barrier onto a specified target location, combining lifting, positioning, and obstacle avoidance. We use the L2-distance to calculate the similarity between observations to judge if the agent has reached the demonstration observation. The threshold for similarity is 0.05. We use 10 demonstrations for training.

\subsection{Adroit}

\begin{figure}[h]
    \centering
    \includegraphics[width=1\textwidth]{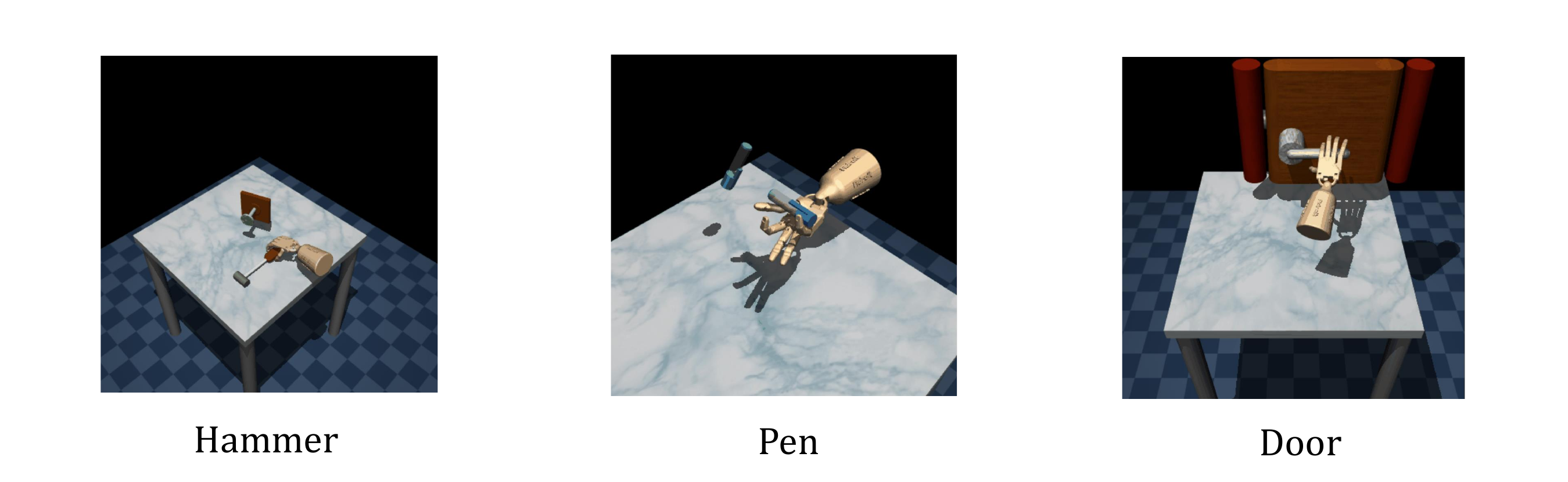}
    \caption{3 environments from Adroit Suite}
    \label{fig:Env-AdroitHand}
\end{figure}

The Adroit Suite~\citep{rajeswaran2017learning} is a set of high-dimensional dexterous hand manipulation tasks that emphasize fine motor control, contact-rich interactions, and sparse-reward learning. It is built upon a 24-DoF ShadowHand robotic hand, presenting significant challenges in both control and generalization. Each task requires the agent to coordinate multiple fingers and joints to manipulate objects with high precision under partial observability and complex dynamics. We use three environments from this suit. Hammer: The agent must grasp a hammer and use it to drive a nail into a box. This task demands stable object manipulation, precise tool orientation, and effective force transmission. Pen: The objective is to reorient and position a pen in an assigned direction. It requires careful control of finger articulation and rotational dexterity. Door: The task involves unlatching and opening a door by manipulating the handle and applying a pulling motion. It tests the agent’s ability to perform multi-stage interactions and coordinate wrist and finger movement to exert torque in the correct direction. We use the Mean square error between images rendered to calculate the similarity to judge if the agent has reached the demonstration observation. Each rendered image will be reshaped to a size (100,100,3). We use 10 demonstrations for training.

\subsection{Maniskill}

\begin{figure}[h]
    \centering
    \includegraphics[width=1\textwidth]{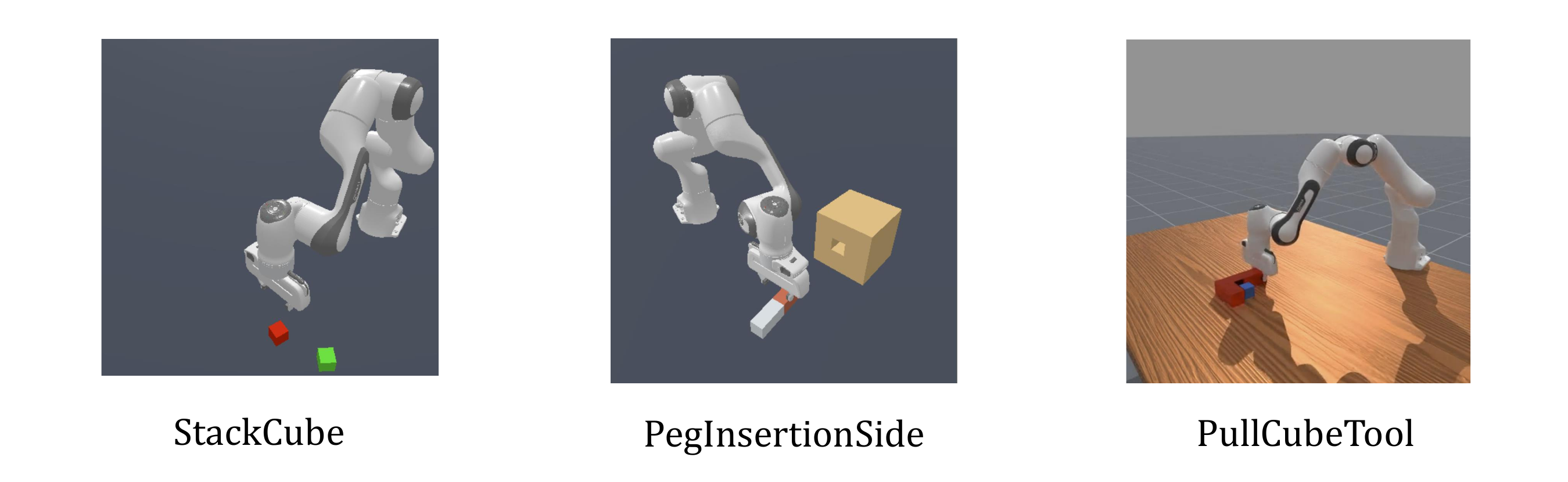}
    \caption{3 environments from Maniskill}
    \label{fig:Env-Maniskill}
\end{figure}

ManiSkill~\citep{gu2023maniskill2, taomaniskill3} is a comprehensive benchmark suite designed to evaluate generalizable robotic manipulation skills in simulation, emphasizing real-world task diversity, object variety, and generalization across instances. It provides high-quality 3D environments with continuous control, supporting both visual and proprioceptive observations. The benchmark is particularly challenging under sparse reward settings, as tasks often require multi-step reasoning, long-horizon planning, and precise control to accomplish. We pick three environments from this benchmark. PullCubeTool: Given an L-shaped tool that is within the reach of the robot, the agent needs to leverage the tool to pull a cube that is out of it’s reach. PegInsertionSide: The robot must align and insert a peg into a side-entry slot. The task demands precise pose estimation, spatial reasoning, and careful control to avoid misalignment or jamming. StackCube: This task involves picking up a cube and accurately stacking it on top of another. We use the Mean square error between images rendered to calculate the similarity to judge if the agent has reached the demonstration observation. Each rendered image will also be reshaped to a size (100,100,3). We use 20 demonstrations for training. On StackCube and PegInsertionSide, we scale up the position(x,y,z) 10 times and normalize the degree of griper opening for more stable learning. We set the clearance of the hole to 0.01 in PegInsertionSide so that the peg could be inserted more easily.

\section{More Experiments}
\label{sec:more-experiments}

\subsection{More Baselines}
\label{subsec:more-baselines}

Generative Adversarial Imitation Learning (\textbf{GAIL})~\citep{ho2016generative, arulkumaran2024pragmatic} adopts an adversarial learning framework where a discriminator is trained to differentiate between expert and agent trajectories; the discriminator's output is then used as the reward signal for the agent. Primal Wasserstein Imitation Learning (\textbf{PWIL})~\citep{dadashi2020primal, arulkumaran2024pragmatic} formulates imitation as a primal optimization problem that minimizes the Wasserstein distance between expert and agent trajectory distributions. It constructs a shaped reward function directly from this distance, encouraging the agent to produce expert-like behaviors. Soft Q Imitation Learning (\textbf{SQIL})~\citep{reddy2019sqil} simplifies imitation learning by assigning fixed rewards to expert and agent transitions, effectively transforming imitation into a standard reinforcement learning problem with sparse binary rewards.
Value-based Distribution Correction Estimation (\textbf{ValueDice})~\citep{kostrikov2019imitation} takes a distribution-matching perspective by minimizing a divergence between expert and agent state-action occupancy measures, providing a principled connection between imitation and value-based reinforcement learning.
Reinforcement Learning with Prior Data (\textbf{RLPD})~\citep{ball2023efficient} is a state-of-the-art baseline improving the efficiency of online reinforcement learning by leveraging offline data. We run these baselines over 5 random experimental seeds and report the average success rate.

\begin{figure}[h]
  \centering
    \subfigure[Disassemble]{\includegraphics[width=0.24\textwidth]{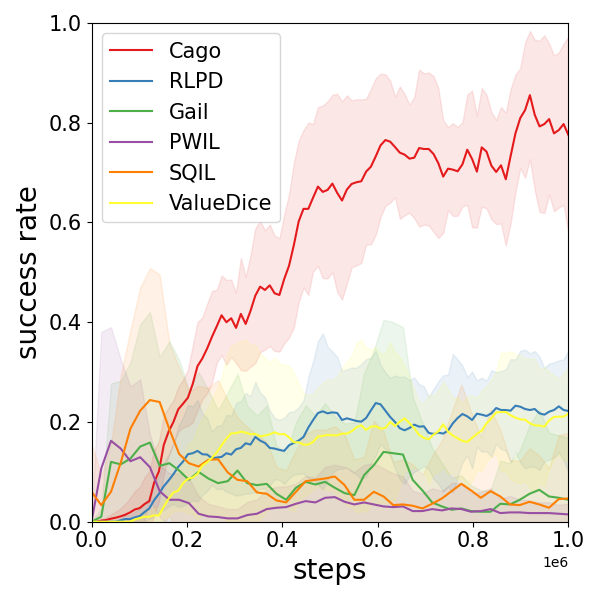}}
    \subfigure[PickPlaceWall]{\includegraphics[width=0.24\textwidth]{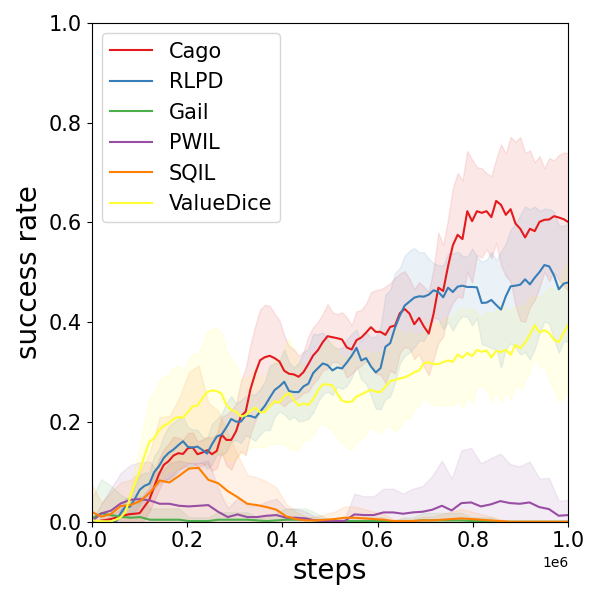}}
    \subfigure[ShelfPlace]{\includegraphics[width=0.24\textwidth]{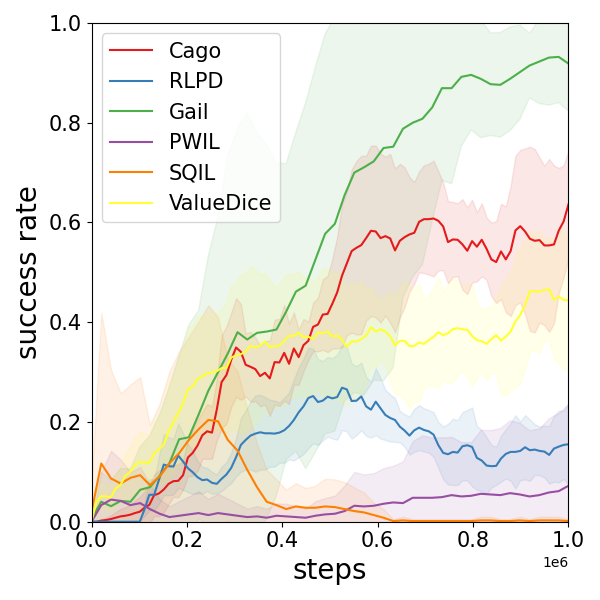}}
    \subfigure[StickPull]{\includegraphics[width=0.24\textwidth]{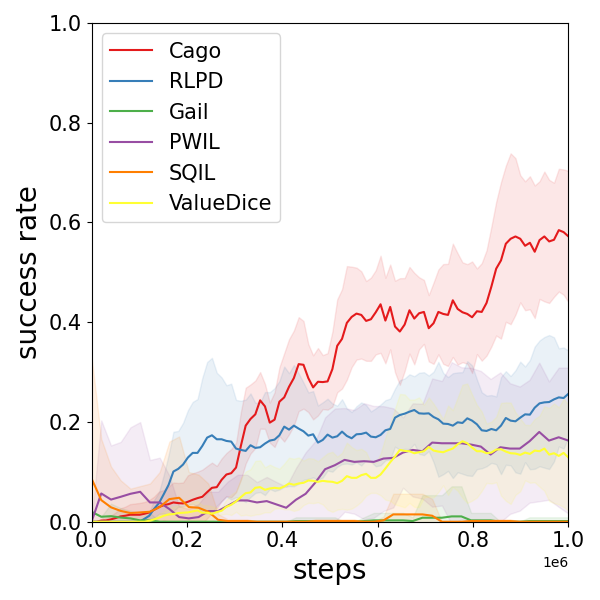}}
    \subfigure[StickPush]{\includegraphics[width=0.24\textwidth]{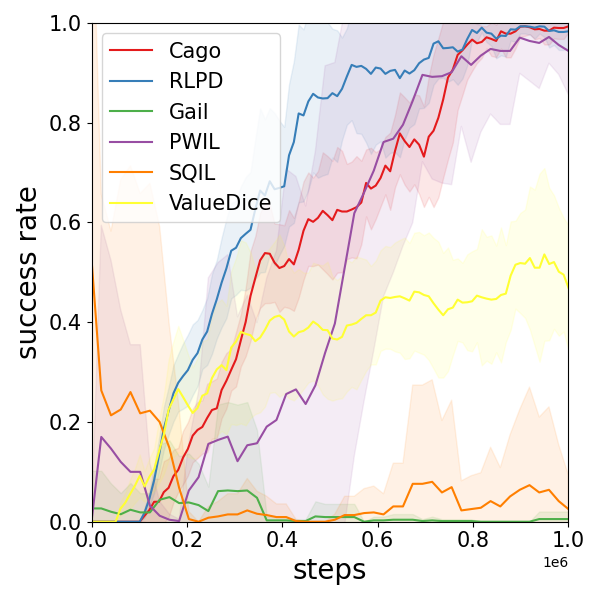}}
    \subfigure[Door]{\includegraphics[width=0.24\textwidth]{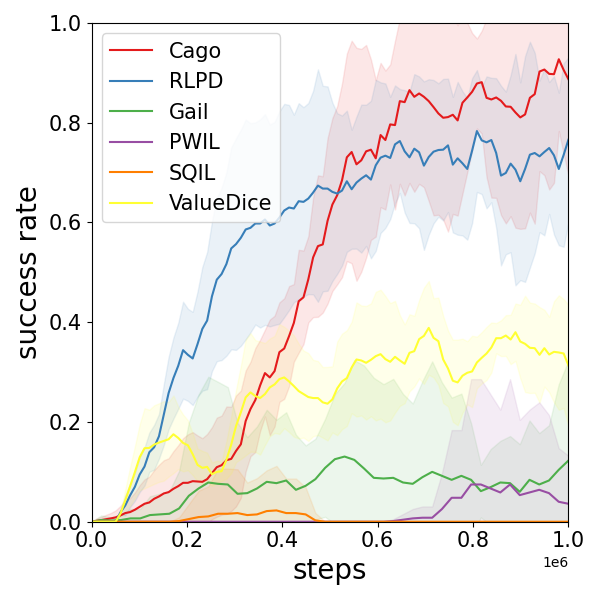}}
    \subfigure[Hammer]{\includegraphics[width=0.24\textwidth]{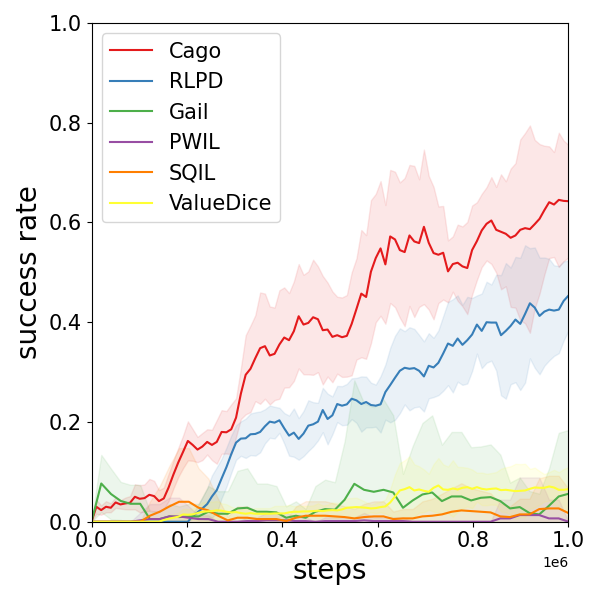}}
    \subfigure[Pen]{\includegraphics[width=0.24\textwidth]{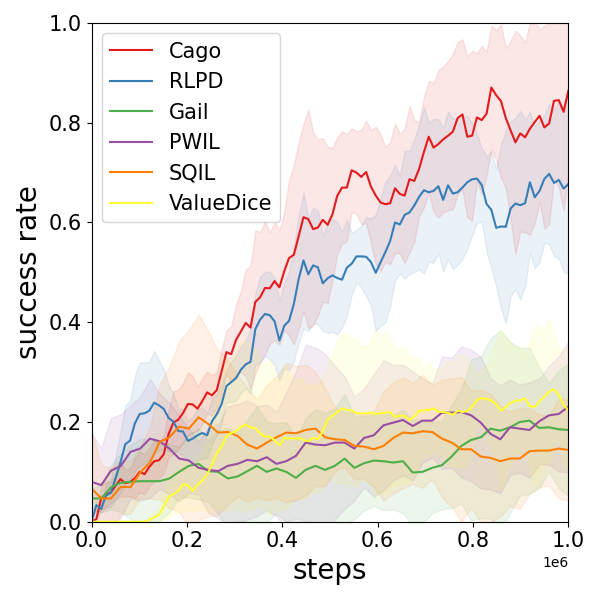}}
    \subfigure[PullCubeTool]{\includegraphics[width=0.24\textwidth]{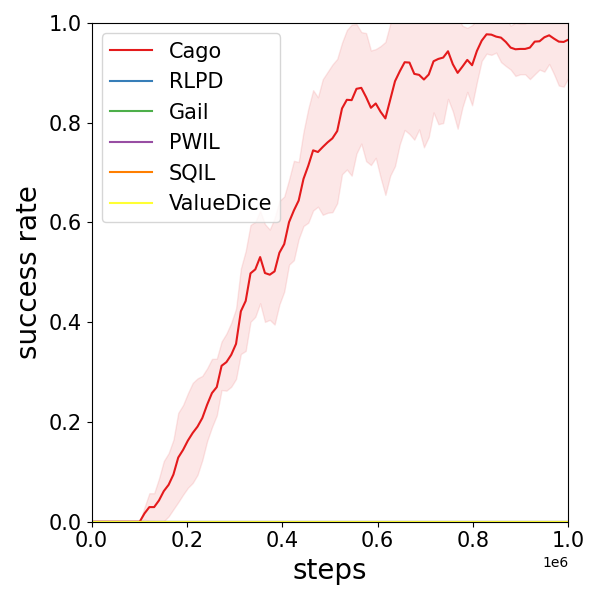}}
    \subfigure[PegInsertionSide]{\includegraphics[width=0.24\textwidth]{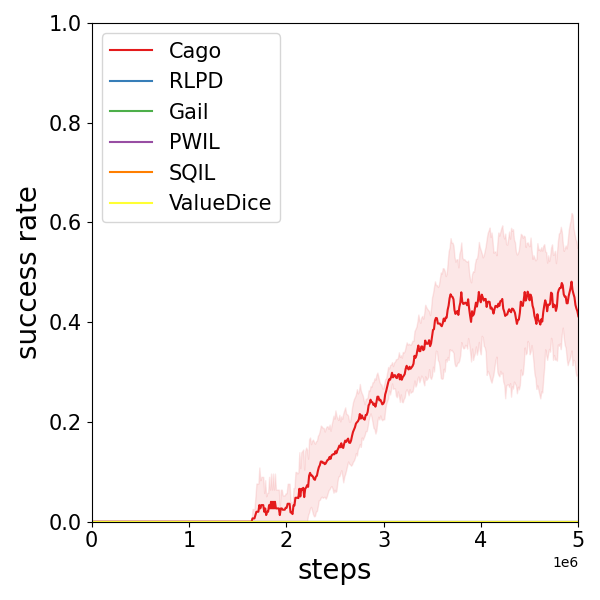}}
    \subfigure[StackCube]{\includegraphics[width=0.24\textwidth]{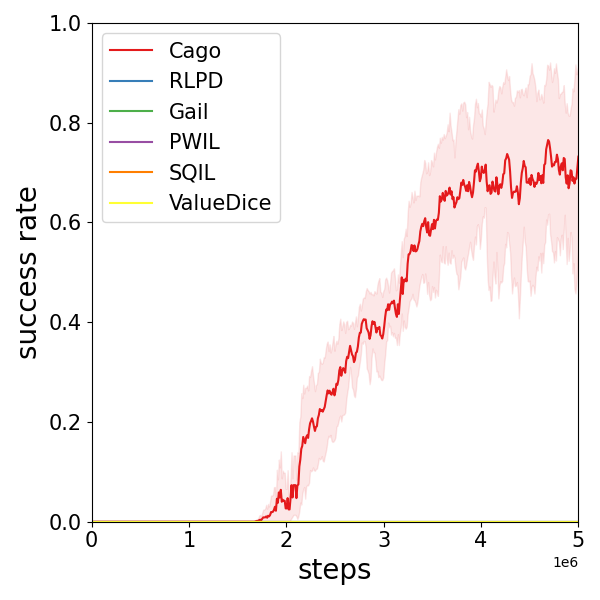}}
  \caption{Experiment results comparing \tool{} with the additional baselines over 5 random seeds.}
  \vspace{-10pt}
  \label{fig:more-exp-results}
\end{figure}

As the results shown in Figure~\ref{fig:more-exp-results}, \tool{} generally outperforms all additional baseline approaches across a diverse set of manipulation tasks. In tasks such as Disassemble, StickPull, Hammer, and Pen, \tool{} demonstrates significantly faster convergence and higher final success rates, indicating its superior learning efficiency and robustness. Particularly in Maniskill environments, \tool{} is the only method that achieves meaningful learning progress, while all baselines fail to get any success, highlighting the importance of capability-aware goal sampling in challenging, sparse-reward environments.
Although GAIL achieves the best performance in the ShelfPlace environment, this success is not representative of its overall effectiveness. In all other tasks, GAIL performs poorly, exhibiting highly unstable learning process and low final success rates. 

\subsection{Visual goal predictor}
\label{subsec: visual-gp}

\begin{figure}[H]
    \centering
    \includegraphics[width=0.8\textwidth]{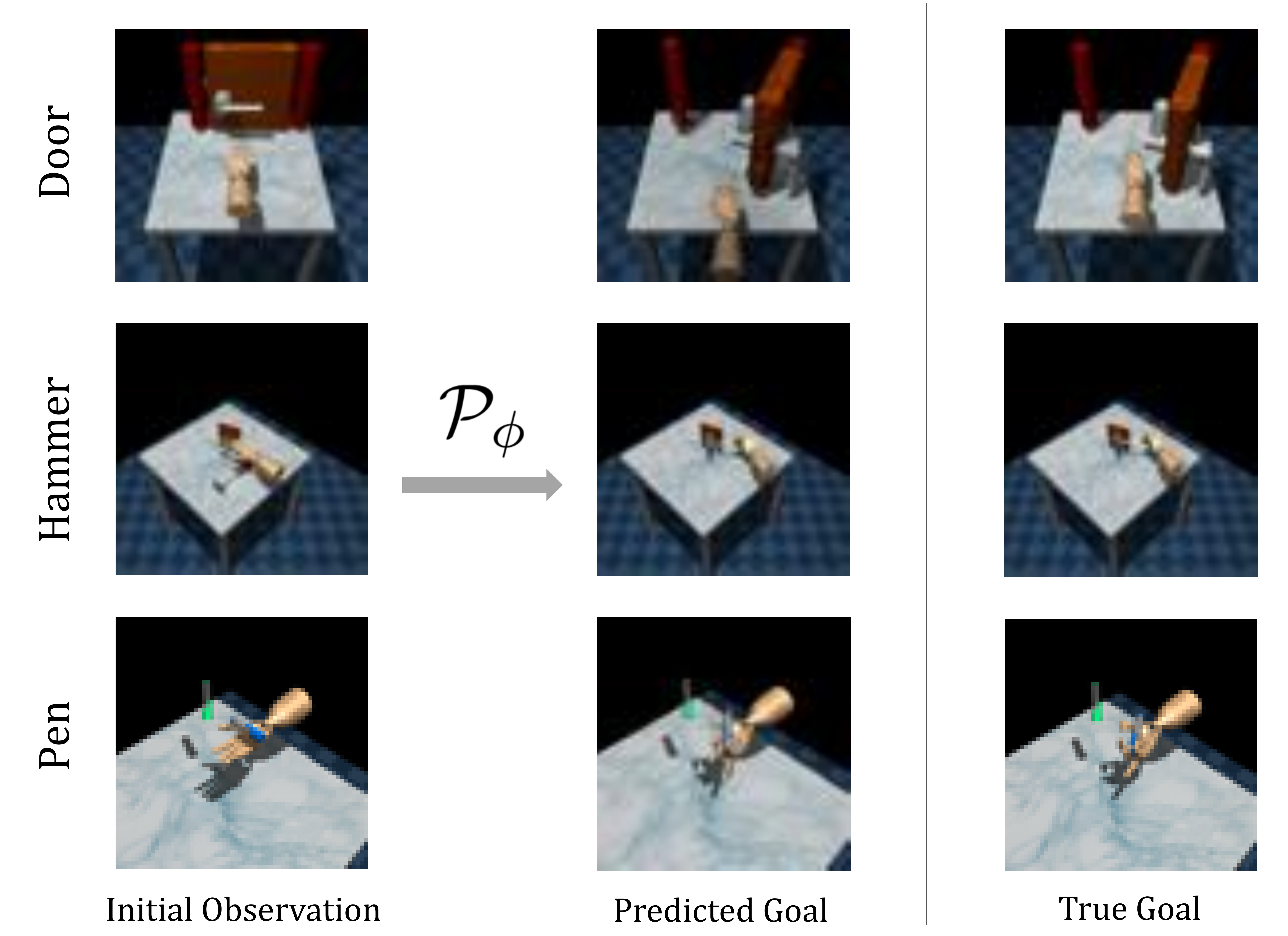}
    \caption{Visual goal prediction results from our learned goal predictor $\mathcal{P}_\phi$. Each row corresponds to a different task (Door, Hammer, Pen). From left to right: the agent's initial observation at test time, the goal image predicted by $\mathcal{P}_\phi$, and the ground-truth final observation from a demonstration trajectory. Notably, these demonstrations are drawn from unseen seeds not used during training. The predicted goals closely match the actual final states, illustrating strong generalization of $\mathcal{P}_\phi$ to novel environment seeds.}
    \vspace{-10pt}
    \label{fig:visual-goal-predicted}
\end{figure}

To evaluate the generalization ability of our visual goal predictor $\mathcal{P}\phi$, we visualize its predicted goal images given initial observations at test time in Figure~\ref{fig:visual-goal-predicted}. For each task, we compare the predicted goal (middle column) to the ground-truth final observation from a demonstration trajectory (right column). Importantly, these demonstrations are collected from unseen seeds that were never used during training.  These results show that $\mathcal{P}\phi$ is capable of accurately inferring the final goal state purely from a single initial image observation, even in unseen evaluation seeds. This predictive capability allows the goal-conditioned policy $\pi^G$ to finish tasks effectively for any environment seeds.

\subsection{Ablation on Number of Demonstration}
\label{subsec: ablation-Num-demo}

This section investigates how the number of demonstration trajectories used for training influences the performance of \tool{} across various tasks. We present success rate curves under different numbers of demonstrations for four representative environments from Metaworld, Adroit, and ManiSkill. The results show that \tool{} maintains strong performance even with a limited amount of expert data (as few as 5 demonstrations), particularly in tasks such as Disassemble, StickPush, and Pen, demonstrating its robustness and stability under data-scarce conditions.
Increasing the number of demonstrations in general yield higher final success rates. This effect is especially evident in more challenging tasks, such as the PegInsertionSide task in Maniskill. These findings suggest that \tool{}'s performance can be further improved with sufficient expert demonstrations. %highlighting its potential for tackling complex tasks when more data is available.

\begin{figure}[H] 
  \centering
    \subfigure[Disassemble]{\includegraphics[width=0.24\textwidth]{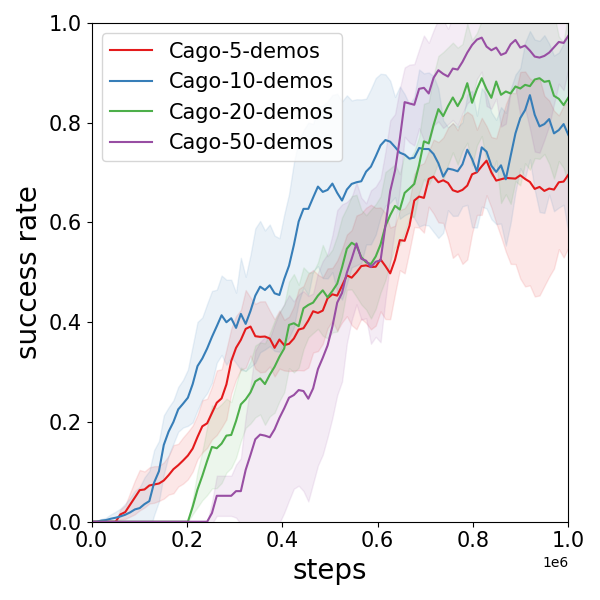}}
    \subfigure[StickPush]{\includegraphics[width=0.24\textwidth]{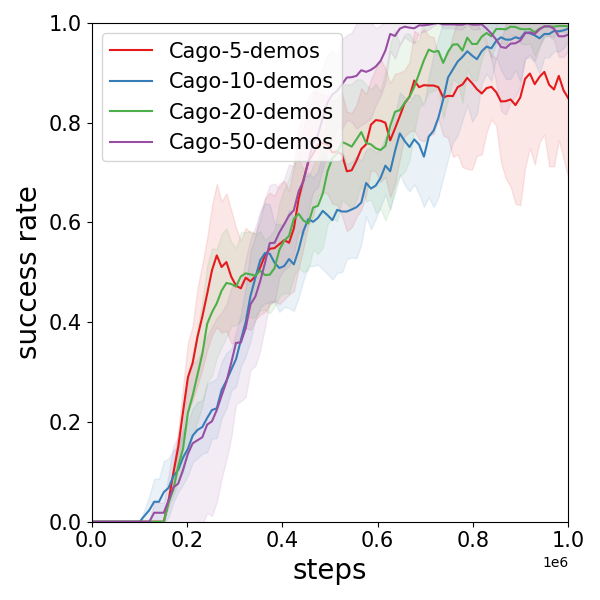}}
    \subfigure[Pen]{\includegraphics[width=0.24\textwidth]{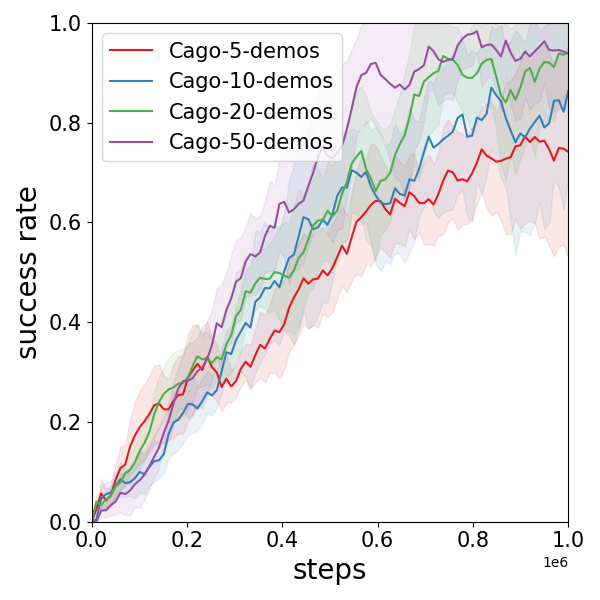}}
    \subfigure[PegInsertionSide]{\includegraphics[width=0.24\textwidth]{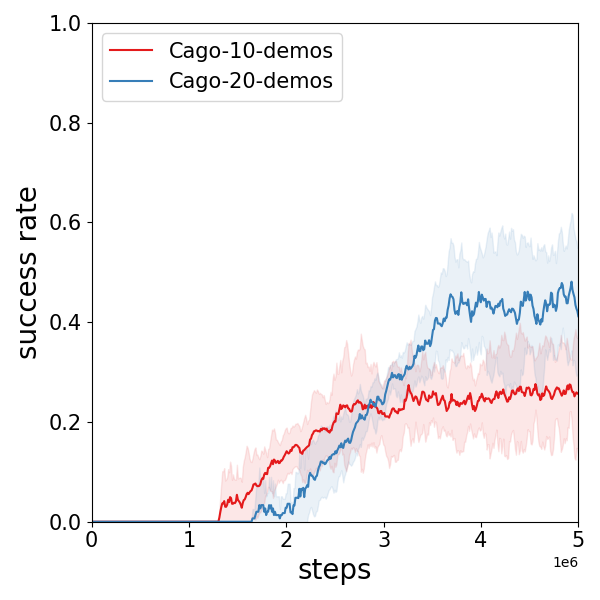}}
  \caption{Success rates under different numbers of demonstrations. Results are averaged over 5 random seeds.}
  \vspace{-10pt}
  \label{fig:num-demo-ablation}
\end{figure}
\subsection{Ablation on Demonstration Quality}
\label{subsec: ablation-subop-demo}

A common concern in demonstration-based reinforcement learning is the reliance on high-quality trajectories that adequately cover the task space. While our theoretical analysis (Theorem~1) assumes such demonstrations in order to establish guarantees, real-world data are often noisy, incomplete, or even contain failed attempts. To evaluate the robustness of \tool{} under such imperfections, we conducted ablation studies on both \emph{suboptimal} and \emph{failed} demonstrations.

\paragraph{Suboptimal Demonstrations.}
We first constructed three types of perturbed demonstrations: 
(i) \emph{Missing Observations}: We randomly removed $20\%$ of the observations from each original demonstration trajectory to simulate incomplete data.
(ii) \emph{Noisy Actions}: We added Gaussian noise (scaled by 0.1) to each action output by the expert policy during expert trajectory collection.
(iii) \emph{Random Actions}: We replaced the expert’s action with a randomly sampled action with a probability of $20\%$ at each timestep during expert trajectory collection.

These settings simulate three kinds of suboptimality: missing state information, noisy expert behavior, and corrupted decision-making. Using these demonstrations, we re-evaluated \tool{} on three benchmark environments: AdroitPen, Stick-Push, and Disassemble. Because we do not have the expert policy for Adroit, we only conducted the "Missing Obs" experiment for Adroit-Pen. Table~\ref{tab:demo-noise} reports the final success rates after 1M training steps, averaged over 5 random seeds (each evaluated over 100 test episodes). The results suggests that \tool{} is robust to the suboptimality in demonstrations. This is because: First, \tool{} only uses demonstration trajectories to extract sequences of observations that serve as curriculum goals. It does not attempt to imitate the demonstration directly, nor does it rely on demonstrations to infer a reward function. This significantly reduces \tool{}'s dependency on high-quality demonstrations. Second, even when the demonstration trajectories are suboptimal—as long as they represent a successful sequence of observations—\tool{} can still learn well. We observe that in the Disassemble environment, \tool{} achieves significantly higher success rates when using the "Random Actions" type of suboptimal demonstrations compared to using the original demonstrations. Those imperfect demonstrations—with pauses or detours—cover a broader region of the task space, thereby improving the final policy's generalization. More importantly, \tool{} is guided by a reward function shaped using a Temporal Distance Network, which estimates how many steps are needed to reach a goal state. This enables \tool{} to discover more efficient, often shorter, alternative paths for goal reaching. Therefore, even when the demonstration are suboptimal, \tool{} may still learn the optimal paths.

\begin{table}[h]
\centering
\caption{Success rates under suboptimal demonstrations (after $1$M steps, averaged over $5$ seeds).}
\vspace{1ex}
\label{tab:demo-noise}
\begin{tabular}{lcccc}
\toprule
Environment & Original Demo & Missing Obs & Noisy Actions & Random Actions \\
\midrule
StickPush   & 0.99 & 0.99 & 0.97 & 0.96 \\
Disassemble  & 0.80 & 0.77 & 0.88 & 0.96 \\
Adroit-Pen   & 0.82 & 0.88 & --   & --   \\
\bottomrule
\end{tabular}
\end{table}

\begin{figure}[H] 
  \centering
    \subfigure[Disassemble]{\includegraphics[width=0.3\textwidth]{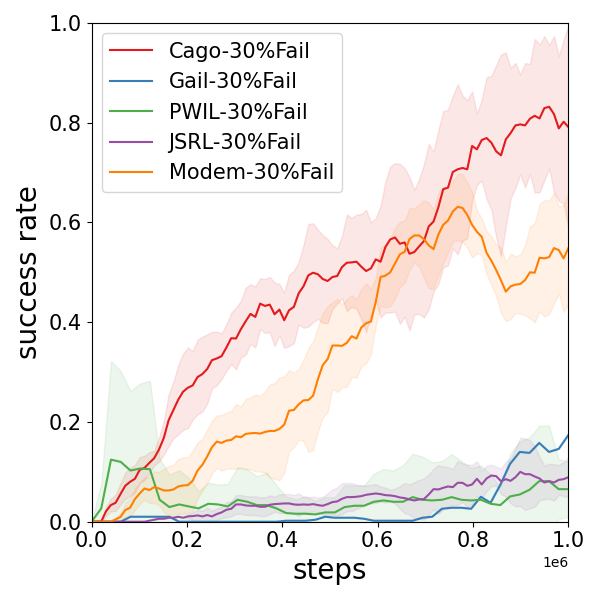}}
    \subfigure[StickPush]{\includegraphics[width=0.3\textwidth]{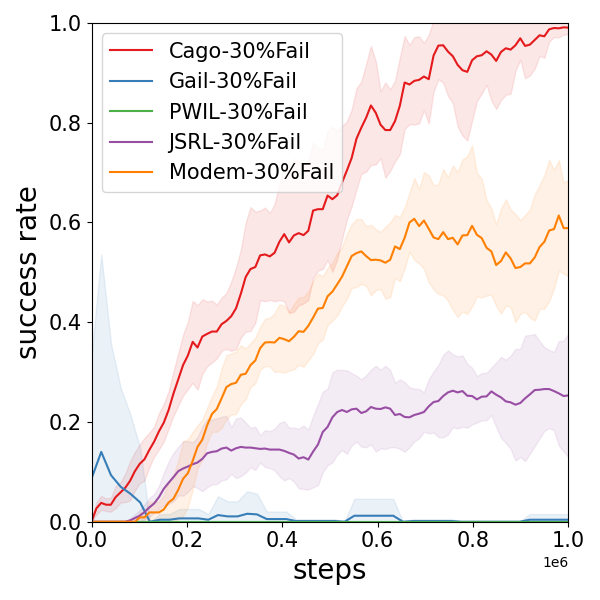}}
    \subfigure[Pen]{\includegraphics[width=0.3\textwidth]{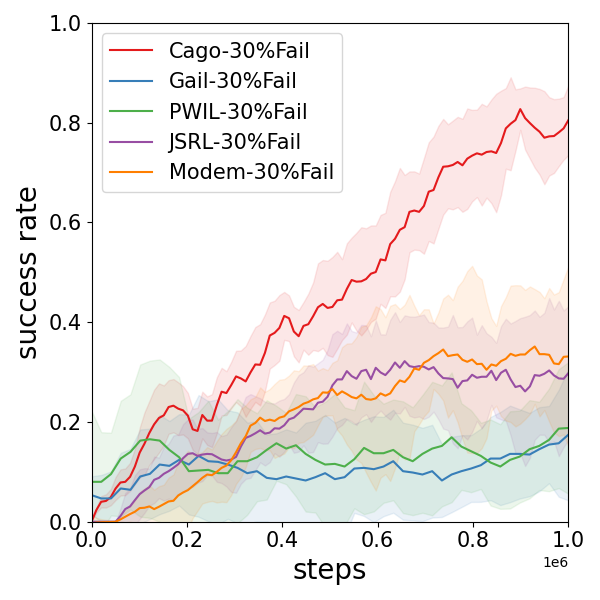}}
  \caption{Success rates under 30\% failed demonstrations (after $1$M steps, averaged over $5$ seeds).}
  \vspace{-10pt}
  \label{fig:30fail}
\end{figure}

\begin{figure}[H] 
  \centering
    \subfigure[Disassemble]{\includegraphics[width=0.3\textwidth]{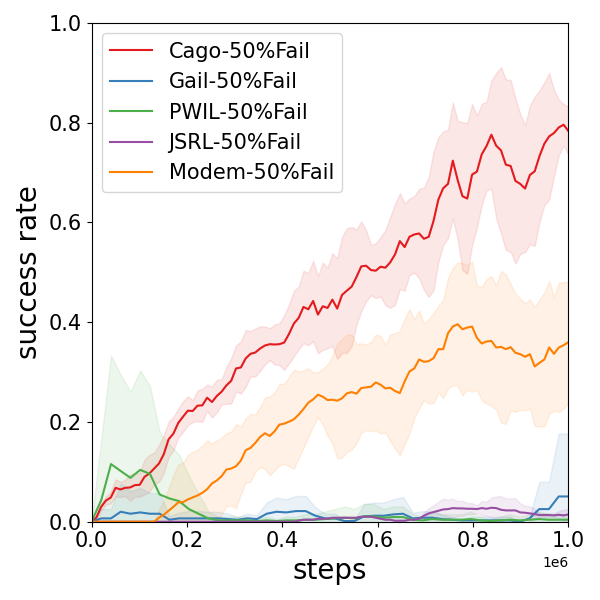}}
    \subfigure[StickPush]{\includegraphics[width=0.3\textwidth]{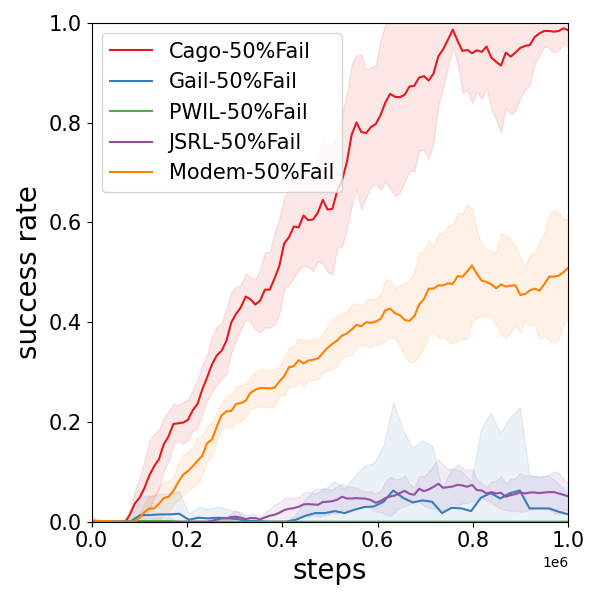}}
    \subfigure[Pen]{\includegraphics[width=0.3\textwidth]{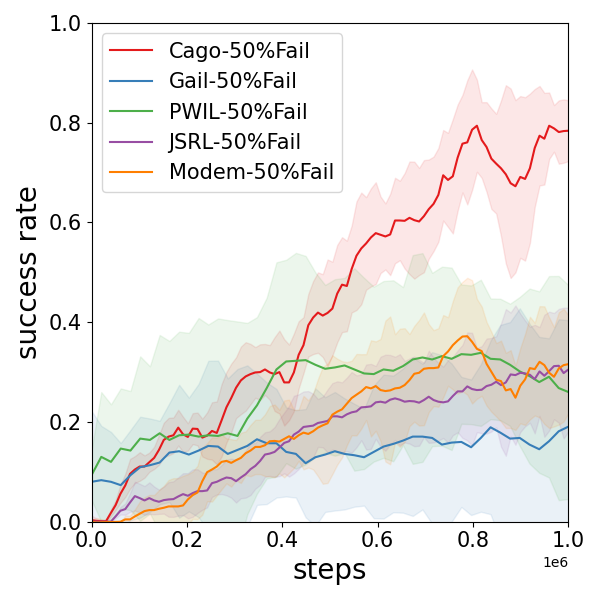}}
  \caption{Success rates under 50\% failed demonstrations (after $1$M steps, averaged over $5$ seeds).}
  \vspace{-10pt}
  \label{fig:50fail}
\end{figure}

\paragraph{Failed Demonstrations.}
We next considered demonstrations that include failed trajectories. For MetaWorld, we injected uniform noise (scale $0.3$) into expert actions, producing datasets with $30\%$ and $50\%$ failed demonstrations. We then compared \tool{} against several strong baselines (GAIL, PWIL, JSRL, and Modem) on Adroit-Pen, Stick-Push, and Disassemble. We assume the environments used to collect demonstrations provide a sparse (binary) reward for goal reaching, enabling us to train goal predictors exclusively on successful demonstrations. The results are shown in Figure~\ref{fig:30fail}, ~\ref{fig:50fail}.

% We compared the performance under such noisy demonstration settings against the original results, in order to assess robustness. The final success rates after 1M training steps, averaged over 5 random seeds are shown in Table~\ref{tab:demo-failed}. The numbers in parentheses refer to the original results reported in the Figure~\ref{fig:exp-results}.

% \begin{table}[h]
% \centering
% \caption{Success rates under failed demonstrations (after $1$M steps, averaged over $5$ seeds).}
% \vspace{1ex}
% \label{tab:demo-failed}
% \begin{tabular}{lccccc}
% \toprule
% \multicolumn{6}{c}{30\% Failed Demonstrations} \\
% \midrule
% Environment & \tool{} (Ours) & GAIL & PWIL & JSRL & Modem \\
% \midrule
% Adroit-Pen   & 0.78 (0.82) & 0.18 (0.17) & 0.26 (0.22) & 0.28 (0.45) & 0.31 (0.32) \\
% Stick-Push   & 0.99 (0.99) & 0.01 (0.01) & 0.74 (0.92) & 0.24 (0.37) & 0.60 (0.75) \\
% Disassemble  & 0.81 (0.80) & 0.06 (0.06) & 0.09 (0.03) & 0.10 (0.19) & 0.54 (0.59) \\
% \midrule
% \multicolumn{6}{c}{50\% Failed Demonstrations} \\
% \midrule
% Adroit-Pen   & 0.76 (0.82) & 0.15 (0.17) & 0.13 (0.22) & 0.32 (0.45) & 0.29 (0.32) \\
% Stick-Push   & 0.98 (0.99) & 0.00 (0.01) & 0.90 (0.92) & 0.04 (0.37) & 0.47 (0.75) \\
% Disassemble  & 0.78 (0.80) & 0.04 (0.06) & 0.00 (0.03) & 0.02 (0.19) & 0.37 (0.59) \\
% \bottomrule
% \end{tabular}
% \end{table}

\tool{} maintains strong performance even with many failed demonstrations, while the baselines degrade significantly. This robustness to suboptimal or noisy data stems from two key factors: (1) \tool{} prioritizes goals that are both challenging and feasible, regardless of whether they originate from successful or failed demonstrations. While failed trajectories may not directly lead to task success, they can expose the agent to diverse regions of the environment, potentially aiding exploration. This allows \tool{} to extract useful learning signals from a broader range of experiences. (2) \tool{} neither imitates demonstrations nor infers reward functions from them. Instead, it uses demonstrations solely to extract sequences of observations that serve as candidate goals. These goals are selected based on the agent’s own learning progress, forming a curriculum that guides policy improvement. This design reduces sensitivity to demonstration quality.

\subsection{Ablation on Parameters}
$\lambda_{visit}$ controls the threshold for how frequently a state must be visited before it is considered “mastered” by the agent, and thus eligible for sampling more difficult goals. If it is set too small, \tool{} may prematurely progress to harder goals before acquiring sufficient competence in easier ones. Conversely, if it is set too large, \tool{} may become overly conservative, spending excessive time on already-mastered states. We have conducted a sensitivity analysis by setting $\lambda_{visit} \in 50,100,200$, and evaluated \tool{} on three environments: Stick-Push, Disassemble, and AdroitPen. The final success rates after 1M training steps, averaged over 5 random seeds are shown in Table~\ref{tab:parameter_lambda_ablation}.

\begin{table}[h]
\centering
\caption{Performance with varying $\lambda_{visit}$.}
\vspace{1ex}
\label{tab:parameter_lambda_ablation}
\begin{tabular}{lccc}
\toprule
Environment & $\lambda_{visit}=50$ & $\lambda_{visit}=100$ & $\lambda_{visit}=200$ \\
\midrule
StickPush       & 0.94 & 0.96 & 0.99 \\
Disassemble     & 0.79 & 0.82 & 0.80 \\
Adroit-Pen      & 0.77 & 0.82 & 0.83 \\
\bottomrule
\end{tabular}
\end{table}

$\delta$ defines the goal sampling window size, i.e., the proportion of expert states from which new goals are sampled during training. A large $\delta$ value may lead to sampling overly distant or inappropriate goals, affecting stability and learning efficiency. In contrast, a very small $\delta$ may overly restrict exploration and hinder curriculum progression. We added a sensitivity experiment comparing the default $\delta = 10\%$ with a smaller value $\delta = 5\%$ and a larger value $\delta = 20\%$, across the same three environments. Table~\ref{tab:parameter_delta_ablation} shows that while larger values can sometimes slightly reduce performance, \tool{} remains capable of learning effective policies in all settings.

\begin{table}[h]
\centering
\caption{Performance with varying $\delta$.}
\vspace{1ex}
\label{tab:parameter_delta_ablation}
\begin{tabular}{lccc}
\toprule
Environment & $\delta=5\%$ & $\delta=10\%$ & $\delta=20\%$ \\
\midrule
StickPush       & 1.00 & 0.99 & 0.99 \\
Disassemble     & 0.81 & 0.80 & 0.75 \\
Adroit-Pen      & 0.76 & 0.82 & 0.81 \\
\bottomrule
\end{tabular}
\end{table}

\section{Runtime}

\subsection{Experiment total runtimes}
\begin{table}[H]
\centering
\caption{Runtimes per experiment.}
\vspace{1ex}
\label{table:runtime}
\begin{tabular}{lccc}
\toprule
Environment       & Runtime (h) & Benchmark    & Steps \\ 
\midrule
ShelfPlace        & 72    & MetaWorld    & 1e6         \\ 
Disassemble       & 72    & MetaWorld    & 1e6         \\ 
StickPull         & 72    & MetaWorld    & 1e6         \\ 
StickPush         & 72    & MetaWorld    & 1e6         \\ 
PickPlaceWall     & 72    & MetaWorld    & 1e6         \\ 
Adroit-Door       & 80    & Adroit       & 1e6         \\ 
Adroit-Hammer     & 85    & Adroit       & 1e6         \\ 
Adroit-Pen        & 83    & Adroit       & 1e6         \\ 
PullCubeTool      & 78    & ManiSkill    & 1e6         \\ 
PegInsertionSide  & 143   & ManiSkill    & 5e6         \\ 
StackCube         & 155   & ManiSkill    & 5e6         \\ 
\bottomrule
\end{tabular}
\end{table}

\subsection{Computation Time for Updating the \tool{} Visitation Record Dictionary}

In this section, we analyze the computational cost associated with updating the visitation record dictionary. Let the length of a sampled trajectory be denoted as $L_{\tau}$, and let $\tau^{(i)}$ represent the demonstration trajectory associated with the same environment reset seed, having length $L_i$. The visitation record dictionary $\text{Dict}_\text{visit}$ is updated according to Equation~\ref{DictUpdate}:
\[
\text{Dict}_{\text{visit}}[\tau^{(i)}][j] \mathrel{+}= 1 \quad \text{if } \text{sim}(s_t, s^{(i)}_j) \leq \epsilon, \quad \forall t \in {1, \dots, L_{\tau}}, \forall j \in {1, \dots, L_i} 
\]

This update rule implies that for each step in the sampled trajectory, a similarity check is performed against all steps in the corresponding demonstration trajectory. Thus, the time required to perform an update of $\text{Dict}_\text{visit}$ can be approximated by:
\[
\text{Time(Update)} \approx \text{Total Steps} \times L_i \times \text{Time(Similarity Calculation)}
\]
The computational cost is therefore influenced by three main factors: the total number of interaction steps, the length of each demonstration trajectory, and the cost of computing the similarity metric. Importantly, the similarity function $\text{sim}(\cdot,\cdot)$ differs by environment, which directly affects computation time. For MetaWorld environments, we utilize L2-distance in the state vector space (i.e., low-dimensional numerical vectors). This calculation is computationally efficient, typically requiring only simple element-wise operations over vector entries. In contrast, for the Adroit and Maniskill environments, the similarity is computed based on MSE in the image space. This involves pixel-wise comparison over image observations, which increases the computational load due to the large input dimensionality (e.g., $100 \times 100 \times 3$).As a result, while the update rule remains structurally the same, the actual runtime overhead for image-based similarity can be substantially higher than that for state-based similarity. The table below summarizes the total runtime, update time, and similarity function used for each environment:

\begin{table}[H]
\centering
\caption{Computation time and similarity function for updating the visitation dictionary $\text{Dict}_\text{visit}$.}
\vspace{1ex}
\label{table:updatetime}
\begin{tabular}{lcccc}
\toprule
Environment       & Steps      & Runtime (h) & Update Time (h) & Similarity\\ 
\midrule
Disassemble       & 1e6        & 72          & 0.05                & State L2\\ 
PickPlaceWall     & 1e6        & 72          & 0.05                & State L2\\ 
ShelfPlace        & 1e6        & 72          & 0.05               & State L2\\ 
StickPull         & 1e6        & 72          & 0.05                & State L2\\ 
StickPush         & 1e6        & 72          & 0.05                & State L2\\ 
Adroit-Door              & 1e6        & 80          & 5.3               & Image MSE\\ 
Adroit-Hammer            & 1e6        & 85          & 5.3             & Image MSE\\ 
Adroit-Pen                & 1e6        & 83          & 5.3               & Image MSE\\ 
PullCubeTool      & 1e6        & 78          & 2.8               & Image MSE\\ 
PegInsertionSide  & 5e6        & 143         & 13.8              & Image MSE\\ 
StackCube         & 5e6        & 155         & 14.2               & Image MSE\\ 
\bottomrule
\end{tabular}
\end{table}

\section{Hyperparameters}
\label{Sec:hyperparameters}

We adopt the default hyperparameters from the LEXA backbone model-based RL (MBRL) agent—such as the learning rate, optimizer, and network architecture—and maintain them consistently across all environments. The primary hyperparameter tuning for \tool{} focuses on the following aspects: (1) the episode length $L_{\tau}$; (2) the proportion of $L_{\tau}$ allocated to the goal-directed phase $T_{go}$; (3) the number of demonstrations $N_{demo}$ used for both dictionary construction and environment resetting; (4) the visit frequency threshold $\lambda_{visit}$ used in Algorithm~\ref{Algorithm-FindGoal} for filtering goal candidates; and (5) the similarity calculate metrics in Equation~\ref{DictUpdate}; (6) the similarity threshold $\epsilon$ in Equation~\ref{DictUpdate}.

\begin{table}[H]
\centering
\caption{Hyperparameters of \tool{}.}
\vspace{1ex}
\label{table:hyperparameters}
\begin{tabular}{lccccccc}
\toprule
Environment       & $L_{\tau} $ &  $T_{go}\ rate$ & $N_{demo}$  & $\lambda_{visit}$  & $\delta$ & $\text{sim}(\cdot,\cdot)$  &  $\epsilon$ \\ 
\midrule
Disassemble                 &  300  &  0.6  &  10  &  200  & 10\% & State L2  & 0.1\\
PickPlaceWall               &  150  &  0.6  &  10  &  200  & 10\% & State L2  & 0.08\\
ShelfPlace                  &  150  &  0.6  &  10  &  50  & 10\% & State L2  & 0.05\\
StickPull                   &  200  &  0.6  &  10  &  200  & 10\% & State L2  & 0.08\\
StickPush                   &  150  &  0.7  &  10  &  200  & 10\% & State L2  & 0.05\\
Adroit-Door                 &  200  &  0.7  &  10  &  100  & 10\% & Image MSE & 200.0\\
Adroit-Hammer               &  100  &  0.6  &  10  &  200  & 10\% & Image MSE & 50.0\\
Adroit-Pen                  &  100  &  0.7  &  10  &  200  & 10\% & Image MSE & 200.0\\
PullCubeTool                &  100  &  0.7  &  10  &  100  & 10\% & Image MSE & 100.0\\
PegInsertionSide            &  100  &  0.7  &  20  &  100  & 10\% & Image MSE & 100.0\\
StackCube                   &  100  &  0.7  &  20  &  100  & 10\% & Image MSE & 50.0\\
\bottomrule
\end{tabular}
\end{table}

% \newpage
% \section{Ablation Experiments on Number of Demonstrations}

% \begin{figure}[H] 
%   \centering
%     \subfigure[Disassemble]{\includegraphics[width=0.24\textwidth]{Exp_results/NumDemo_Metaworld_Disassemble_success_rate.png}}
%     \subfigure[StickPush]{\includegraphics[width=0.24\textwidth]{Exp_results/NumDemo_Metaworld_StickPush_success_rate.png}}
%     \subfigure[Pen]{\includegraphics[width=0.24\textwidth]{Exp_results/NumDemo_Adroit_Pen_success_rate.png}}
%     \subfigure[PegInsertionSide]{\includegraphics[width=0.24\textwidth]{Exp_results/NumDemo_Maniskill_PegInsertion_success_rate.png}}
%   \caption{.}
%   \vspace{-10pt}
%   \label{fig:num-demo-ablation}
% \end{figure}

%%%%%%%%%%%%%%%%%%%%%%%%%%%%%%%%%%%%%%%%%%%%%%%%%%%%%%%%%%%%

\newpage
\section*{NeurIPS Paper Checklist}

%%% BEGIN INSTRUCTIONS %%%
The checklist is designed to encourage best practices for responsible machine learning research, addressing issues of reproducibility, transparency, research ethics, and societal impact. Do not remove the checklist: {\bf The papers not including the checklist will be desk rejected.} The checklist should follow the references and follow the (optional) supplemental material.  The checklist does NOT count towards the page
limit. 

Please read the checklist guidelines carefully for information on how to answer these questions. For each question in the checklist:
\begin{itemize}
    \item You should answer \answerYes{}, \answerNo{}, or \answerNA{}.
    \item \answerNA{} means either that the question is Not Applicable for that particular paper or the relevant information is Not Available.
    \item Please provide a short (1–2 sentence) justification right after your answer (even for NA). 
   % \item {\bf The papers not including the checklist will be desk rejected.}
\end{itemize}

{\bf The checklist answers are an integral part of your paper submission.} They are visible to the reviewers, area chairs, senior area chairs, and ethics reviewers. You will be asked to also include it (after eventual revisions) with the final version of your paper, and its final version will be published with the paper.

The reviewers of your paper will be asked to use the checklist as one of the factors in their evaluation. While "\answerYes{}" is generally preferable to "\answerNo{}", it is perfectly acceptable to answer "\answerNo{}" provided a proper justification is given (e.g., "error bars are not reported because it would be too computationally expensive" or "we were unable to find the license for the dataset we used"). In general, answering "\answerNo{}" or "\answerNA{}" is not grounds for rejection. While the questions are phrased in a binary way, we acknowledge that the true answer is often more nuanced, so please just use your best judgment and write a justification to elaborate. All supporting evidence can appear either in the main paper or the supplemental material, provided in appendix. If you answer \answerYes{} to a question, in the justification please point to the section(s) where related material for the question can be found.

IMPORTANT, please:
\begin{itemize}
    \item {\bf Delete this instruction block, but keep the section heading ``NeurIPS Paper Checklist"},
    \item  {\bf Keep the checklist subsection headings, questions/answers and guidelines below.}
    \item {\bf Do not modify the questions and only use the provided macros for your answers}.
\end{itemize}

%%% END INSTRUCTIONS %%%

\begin{enumerate}

\item {\bf Claims}
    \item[] Question: Do the main claims made in the abstract and introduction accurately reflect the paper's contributions and scope?
    \item[] Answer: \answerYes{} % Replace by \answerYes{}, \answerNo{}, or \answerNA{}.
    \item[] Justification: Our abstract and introduction accurately represent the primary contributions of the paper, which include the development of the \tool{} algorithm designed to solve sparse-reward tasks through a capability-aware goal picking strategy. The introduction outlines the key challenges in GCRL, specifically with sparse rewards, and how \tool{} addresses these by improving the quality of online sampled trajectories using demonstration. These claims are well-supported by the theoretical underpinnings and experimental results presented in the paper, reflecting the scope and impact of the proposed method. 

\item {\bf Limitations}
    \item[] Question: Does the paper discuss the limitations of the work performed by the authors?
    \item[] Answer: \answerYes{} % Replace by \answerYes{}, \answerNo{}, or \answerNA{}.
    \item[] Justification: Our paper thoroughly discusses the limitations of the \tool{} framework in Appendix.

\item {\bf Theory assumptions and proofs}
    \item[] Question: For each theoretical result, does the paper provide the full set of assumptions and a complete (and correct) proof?
    \item[] Answer: \answerYes{} % Replace by \answerYes{}, \answerNo{}, or \answerNA{}.
    \item[] Justification: Yes, our paper provides the full set of assumptions and a complete (and correct) proof in the Appendix~\ref{sec:proof} for Theorem~\ref{theorem}.

    \item {\bf Experimental result reproducibility}
    \item[] Question: Does the paper fully disclose all the information needed to reproduce the main experimental results of the paper to the extent that it affects the main claims and/or conclusions of the paper (regardless of whether the code and data are provided or not)?
    \item[] Answer: \answerYes{} % Replace by \answerYes{}, \answerNo{}, or \answerNA{}.
    \item[] Justification: In the Experiment section and appendix of our paper, we elaborate on the procedure and configuration of our experiments. 
    This includes the sources and modifications of all testing environments, and implementation methods for all baselines, the devices and memory utilized, as well as specific values of hyperparameters employed. Concurrently, we have open-sourced our code; please refer to the Reproducibility Statement section for further details.

\item {\bf Open access to data and code}
    \item[] Question: Does the paper provide open access to the data and code, with sufficient instructions to faithfully reproduce the main experimental results, as described in supplemental material?
    \item[] Answer: \answerYes{} % Replace by \answerYes{}, \answerNo{}, or \answerNA{}.
    \item[] Justification: As we mentioned in the previous justification, we have not only open-sourced our code but also provided detailed steps and settings for reproducing our main experimental results. 
    In the Experiment section and Appendix, we elaborate on the sources and modifications of the environments, baseline implementation details, and \tool{} implementation specifics.

\item {\bf Experimental setting/details}
    \item[] Question: Does the paper specify all the training and test details (e.g., data splits, hyperparameters, how they were chosen, type of optimizer, etc.) necessary to understand the results?
    \item[] Answer: \answerYes{} % Replace by \answerYes{}, \answerNo{}, or \answerNA{}.
    \item[] Justification: We provide comprehensive details regarding the hyperparameters essential for understanding the experiments, including those specific to our \tool{} framework. 
    The table presented (Table.~\ref{table:hyperparameters}) outlines these hyperparameters for each task, facilitating reproducibility and comparison.

\item {\bf Experiment statistical significance}
    \item[] Question: Does the paper report error bars suitably and correctly defined or other appropriate information about the statistical significance of the experiments?
    \item[] Answer: \answerYes{} % Replace by \answerYes{}, \answerNo{}, or \answerNA{}.
    \item[] Justification: We conducted each experiment a minimum of five times using different random seeds, and upon plotting the results, as demonstrated in the Experiment section, we incorporated the experimental error. The solid line denotes the average success rate, while the shaded region signifies the standard deviation among the repeated experimental outcomes.

\item {\bf Experiments compute resources}
    \item[] Question: For each experiment, does the paper provide sufficient information on the computer resources (type of compute workers, memory, time of execution) needed to reproduce the experiments?
    \item[] Answer: \answerYes{} % Replace by \answerYes{}, \answerNo{}, or \answerNA{}.
    \item[] Justification: We clearly specifies the computer resources (8 Nvidia A100 GPU) and the amount of GPU memory required (approximately 2.4GB). 
    
\item {\bf Code of ethics}
    \item[] Question: Does the research conducted in the paper conform, in every respect, with the NeurIPS Code of Ethics \url{https://neurips.cc/public/EthicsGuidelines}?
    \item[] Answer: \answerYes{} % Replace by \answerYes{}, \answerNo{}, or \answerNA{}.
    \item[] Justification: The research conducted in our paper aligns with the NeurIPS Code of Ethics. We have thoroughly reviewed the guidelines and ensured that our research adheres to ethical standards. Additionally, we have implemented measures to safeguard anonymity and comply with pertinent laws and regulations.

\item {\bf Broader impacts}
    \item[] Question: Does the paper discuss both potential positive societal impacts and negative societal impacts of the work performed?
    \item[] Answer: \answerNA{} % Replace by \answerYes{}, \answerNo{}, or \answerNA{}.
    \item[] Justification: Our research aims to address the learning efficiency problem in Reinforcement Learning (RL) within the GCRL environment. It is currently in the theoretical research stage and has minimal societal impact.
    
\item {\bf Safeguards}
    \item[] Question: Does the paper describe safeguards that have been put in place for responsible release of data or models that have a high risk for misuse (e.g., pretrained language models, image generators, or scraped datasets)?
    \item[] Answer: \answerNA{} % Replace by \answerYes{}, \answerNo{}, or \answerNA{}.
    \item[] Justification: Our paper poses no such risks. 

\item {\bf Licenses for existing assets}
    \item[] Question: Are the creators or original owners of assets (e.g., code, data, models), used in the paper, properly credited and are the license and terms of use explicitly mentioned and properly respected?
    \item[] Answer: \answerYes{} % Replace by \answerYes{}, \answerNo{}, or \answerNA{}.
    \item[] Justification: Our paper properly credits the creators or original owners of assets used, including code, data, and models. 

\item {\bf New assets}
    \item[] Question: Are new assets introduced in the paper well documented and is the documentation provided alongside the assets?
    \item[] Answer: \answerYes{} % Replace by \answerYes{}, \answerNo{}, or \answerNA{}.
    \item[] Justification: We have documented our code and provided detailed instructions on its usage, licenses, and permissible scope of use. 
    Additionally, we have included the documentation alongside the assets to ensure accessibility and clarity for users.

\item {\bf Crowdsourcing and research with human subjects}
    \item[] Question: For crowdsourcing experiments and research with human subjects, does the paper include the full text of instructions given to participants and screenshots, if applicable, as well as details about compensation (if any)? 
    \item[] Answer: \answerNA{} % Replace by \answerYes{}, \answerNo{}, or \answerNA{}.
    \item[] Justification: Our paper not involve crowdsourcing nor research with human subjects.

\item {\bf Institutional review board (IRB) approvals or equivalent for research with human subjects}
    \item[] Question: Does the paper describe potential risks incurred by study participants, whether such risks were disclosed to the subjects, and whether Institutional Review Board (IRB) approvals (or an equivalent approval/review based on the requirements of your country or institution) were obtained?
    \item[] Answer: \answerNA{} % Replace by \answerYes{}, \answerNo{}, or \answerNA{}.
    \item[] Justification: Our paper does not involve crowdsourcing nor research with human subjects.

\item {\bf Declaration of LLM usage}
    \item[] Question: Does the paper describe the usage of LLMs if it is an important, original, or non-standard component of the core methods in this research? Note that if the LLM is used only for writing, editing, or formatting purposes and does not impact the core methodology, scientific rigorousness, or originality of the research, declaration is not required.
    %this research? 
    \item[] Answer: \answerNA{} % Replace by \answerYes{}, \answerNo{}, or \answerNA{}.
    \item[] Justification: Our method does not involve LLMs as any important, original, or non-standard components.

\end{enumerate}

\end{document}